%% file: New_IEEEtran_how-to.tex
\documentclass[lettersize,journal,11pt,nofonttune]{IEEEtran}
\usepackage{amsmath,amsfonts}
\usepackage{algorithmic}
\usepackage{array}
\usepackage{textcomp}
\usepackage{stfloats}
\usepackage{url}
\usepackage{verbatim}
\usepackage{graphicx}
\usepackage{xspace}
\usepackage{amsmath}
\usepackage{svg}
\def\BibTeX{{\rm B\kern-.05em{\sc i\kern-.025em b}\kern-.08em
    T\kern-.1667em\lower.7ex\hbox{E}\kern-.125emX}}
\usepackage{balance}
\usepackage[acronyms,nonumberlist,nopostdot,nomain,nogroupskip,acronymlists={hidden}]{glossaries}
\usepackage{soul}
\usepackage{color}
\newglossary[algh]{hidden}{acrh}{acnh}{Hidden Acronyms}
\newcommand{\framework}{AgentRAN\xspace}
\usepackage{tikz}
\usepackage{pgfplots}
\pgfplotsset{compat=1.18}  
\newlength\fheight
\newlength\fwidth
\input{acronyms}
\usepackage{booktabs}
\usepackage{balance}

\usepackage{enumitem}
\setlist[itemize]{leftmargin=*}

\usepackage[footnotesize]{caption}
\usepackage{subcaption}
\usepackage[subtle]{savetrees}

\usepackage[many]{tcolorbox}
\usepackage{lipsum}

\usepackage{dialogue}
\renewenvironment{dialogue}{%
  \begin{list}{}{\setlength{\leftmargin}{0pt}%
  \setlength{\rightmargin}{0pt}%
  \setlength{\listparindent}{0pt}%
  \setlength{\itemindent}{0pt}%
  \setlength{\topsep}{0pt}%
  \setlength{\parsep}{0pt}}%
  \item[]%
}{%
  \end{list}%
}

\newcommand{\speak}[2]{%
  \footnotesize
  \item \textsc{#1:} #2%
}

\bstctlcite{IEEEexample:BSTcontrol}

\usepackage{xcolor}
\definecolor{l2color}{RGB}{0,102,204}
\definecolor{ulcolor}{RGB}{204,102,0}
\definecolor{dlcolor}{RGB}{102,153,0}
\definecolor{pccolor}{RGB}{153,0,153}
\definecolor{dialoguebg}{RGB}{0,128,128}

\newtcolorbox{llmdialogue}[1][]{
  breakable,
  enhanced,
  arc=0pt,
  outer arc=0pt,
  colframe=dialoguebg,
  colback=dialoguebg!05,
  overlay unbroken and first={
    \node[
      draw=dialoguebg,
      fill=dialoguebg,
      rotate=0,
      anchor=north west,
      text=white,
      font=\footnotesize\bfseries
    ]
    at (frame.north west)  
    {#1};
  }
}

\newcommand{\nrric}{Near-Real-Time \gls{ric}\xspace}
\newcommand{\nonric}{Non-Real-Time \gls{ric}\xspace}
\usepackage{titlesec}
\titlespacing*{\subsection}{0pt}{8pt}{6pt}
\titlespacing*{\section}{0pt}{8pt}{6pt}

\newif\ifshowchanges
\newif\ifshowremoved

\newif\ifshowchanges
\newif\ifshowremoved

\showchangestrue
\showremovedfalse

\newcommand{\rev}[1]{\ifshowchanges{#1}\fi}
\newcommand{\removed}[1]{\ifshowremoved{\color{red}#1}\fi}

\begin{document}
\bstctlcite{IEEEexample:BSTcontrol}
\setlength{\floatsep}{3pt} 
\setlength{\textfloatsep}{3pt} 
\setlength{\intextsep}{3pt} 
\setlength{\abovecaptionskip}{0pt} 
\setlength{\belowcaptionskip}{0pt} 
\IEEEaftertitletext{\vspace{-3.\baselineskip}}
\setlength{\dblfloatsep}{8pt} 
\setlength{\dbltextfloatsep}{8pt} 

\title{AgentRAN: An Agentic AI Architecture for Autonomous Control of Open 6G Networks

\thanks{$^\dagger$Institute for Intelligent Networked Systems, \gls{neu}, Boston, MA, USA. E-mail: \{m.elkael, s.doro, l.bonati, m.polese, melodia\}@northeastern.edu. $^\ddagger$\gls{sb}, Japan. E-mail: yunseong.lee.1992@ieee.org, koichiro.furueda@g.softbank.co.jp}
\thanks{This work is partially supported by the NTIA's Public Wireless Supply Chain Innovation Fund (PWSCIF) under Award No. 25-60-IF054 and by OUSD(R\&E) through Army Research Laboratory Cooperative Agreement Number W911NF-24-2-0065. The views and conclusions contained in this document are those of the authors and should not be interpreted as representing the official policies, either expressed or implied, of the Army Research Laboratory or the U.S. Government. The U.S. Government is authorized to reproduce and distribute reprints for Government purposes notwithstanding any copyright notation herein. This work is also partially supported by the U.S. NSF under award CNS-2112471 and by SoftBank Corp.\vspace{-10pt}}}

\author{Maxime Elkael$^\dagger$, Salvatore D'Oro$^\dagger$, Leonardo Bonati$^\dagger$, \\Michele Polese$^\dagger$, Yunseong Lee$^\ddagger$, Koichiro Furueda$^\ddagger$, Tommaso Melodia$^\dagger$}


\maketitle

\begin{abstract}
Despite the programmable architecture of Open \gls{ran}, today's deployments still rely heavily on static control and manual operations.
To move beyond this limitation, we introduce \framework, an AI-native, Open RAN-aligned agentic framework that generates and orchestrates a fabric of distributed \gls{ai} agents based on \gls{nl} intents. Unlike traditional approaches that require explicit programming, \framework's \gls{llm}-powered agents interpret natural language intents, negotiate strategies through structured conversations, and orchestrate control loops across the network. \framework instantiates a self-organizing hierarchy of agents that decompose complex intents across time scales (from sub-millisecond to minutes), spatial domains (cell to network-wide), and protocol layers (PHY/MAC to RRC). 
\removed{A central innovation is the ``\gls{ai}-\gls{ran} Factory'', an automated synthesis pipeline that observes agent interactions and continuously generates new agents embedding improved control algorithms, effectively transforming the network from a static collection of functions into an adaptive system capable of evolving its own intelligence. 
We demonstrate \framework through live experiments on 5G testbeds where competing user demands are dynamically balanced through cascading intents. By replacing rigid APIs with \gls{nl} coordination, \framework fundamentally redefines how future 6G networks autonomously interpret, adapt, and optimize their behavior to meet operator goals.}
\rev{A central innovation is the \gls{ai}-\gls{ran} Factory, which continuously generates improved agents and algorithms from operational data, transforming the network into a system that evolves its own intelligence.
We validate \framework through live 5G experiments, demonstrating dynamic adaptation to changing operator intents across power control and scheduling. Key benefits include transparent decision-making (all agent reasoning is auditable), bootstrapped intelligence (no initial training data required), and continuous self-improvement via the AI-RAN Factory.}
\end{abstract}

\glsresetall

\glsunset{gpu} 
\glsunset{cpu} 
\glsunset{ram}
\glsunset{oran}
\glsunset{3gpp}
\glsunset{xapp}
\glsunset{rapp}
\glsunset{dapp}

\section{Introduction} \label{sec:intro}

As cellular systems evolve toward \gls{6g}~\rev{\cite{na2024operator}}, network architectures adopt disaggregated, software-based approaches thanks to standardization and development efforts from 3GPP and the O-RAN ALLIANCE. 
\removed{As these efforts push the \gls{ran} toward open and programmable systems, they have yet to achieve the level of flexibility, autonomy, and intelligence achieved by large-scale cloud deployments, where capabilities such as dynamic workload management through auto-scaling~\cite{chen2018survey}, serverless computing~\cite{shafiei2022serverless}, and intent-based orchestration—including emerging agentic \gls{ai} paradigms~\cite{rao2024eco}- are already commonplace.}
\rev{However, a significant gap remains: while cloud platforms routinely employ autonomous orchestration, auto-scaling, and intent-based management~\cite{chen2018survey,shafiei2022serverless,rao2024eco}, cellular networks still rely heavily on static configurations and manual operations. Any change in network conditions (e.g., traffic surges, emergency events, shifting user priorities) requires manual reconfiguration of parameters across multiple network elements, a slow and error-prone process.}

\rev{\textbf{The Challenge.} Consider a simple scenario: an operator wants to prioritize emergency sensor traffic over regular broadband users. Today, this requires manually adjusting scheduling weights, power control targets, and resource allocation across potentially dozens of parameters and network elements. What if, instead, the operator could simply express this intent in \gls{nl}, and the network would autonomously understand how to achieve it?}
\removed{To bridge this gap, we argue that realizing an AI-native \gls{ran} demands rethinking the very notion of a network control element.} \rev{This is the vision behind \framework: a hierarchy of \textit{AI agents}---autonomous software entities with four key agentic properties: \textit{perception} (monitor network conditions and KPIs), \textit{memory} (store context, historical data and past decisions), \textit{reasoning} (use \glspl{llm} to interpret intents and plan decisions), and \textit{action} (adjust parameters or create new agents via the AI-RAN Factory). These agents coordinate through \gls{nl}, making every decision transparent and auditable by humans.}
\removed{Programmability and \gls{ai} are seen as enablers of \gls{ibn} in cellular 
networks, translating high-level operator directives into automated \gls{ran} 
behaviors~\cite{leivadeas2023survey}. Some early results on intent-driven 
\gls{ra} are discussed in~\cite{nahum2024intent,elkael2025allstar}. 
Recent work on symbiotic agents~\cite{chatzistefanidis2025symbioticagentsnovelparadigm} 
shows that one can pair \glspl{llm} with 
optimizers (e.g., PID controllers) to improve decision accuracy. Despite making a step in the right direction, these works remain 
limited to single-variable negotiations and static agent architectures, 
lacking the ability to coordinate heterogeneous control variables across 
domains or autonomously generate new agents as network conditions evolve. 
These factors, combined with tight compute requirements at the physical layer, demanding \glspl{sla}, strict regulations and network complexity have led to risk adversity with a preference for closed, monolithic and single-vendor solutions, which slows down adoption of cloud-native technologies in cellular networks.}
\rev{Recent work has explored intent-driven resource allocation~\cite{nahum2024intent,elkael2025allstar} and pairing \glspl{llm} with classical optimizers~\cite{chatzistefanidis2025symbioticagentsnovelparadigm}. However, these approaches are limited to single control variables and static architectures, lacking coordination across domains or the ability to generate new agents as conditions evolve.}

\begin{figure*}
    \centering
    \includegraphics[width=.8\linewidth]{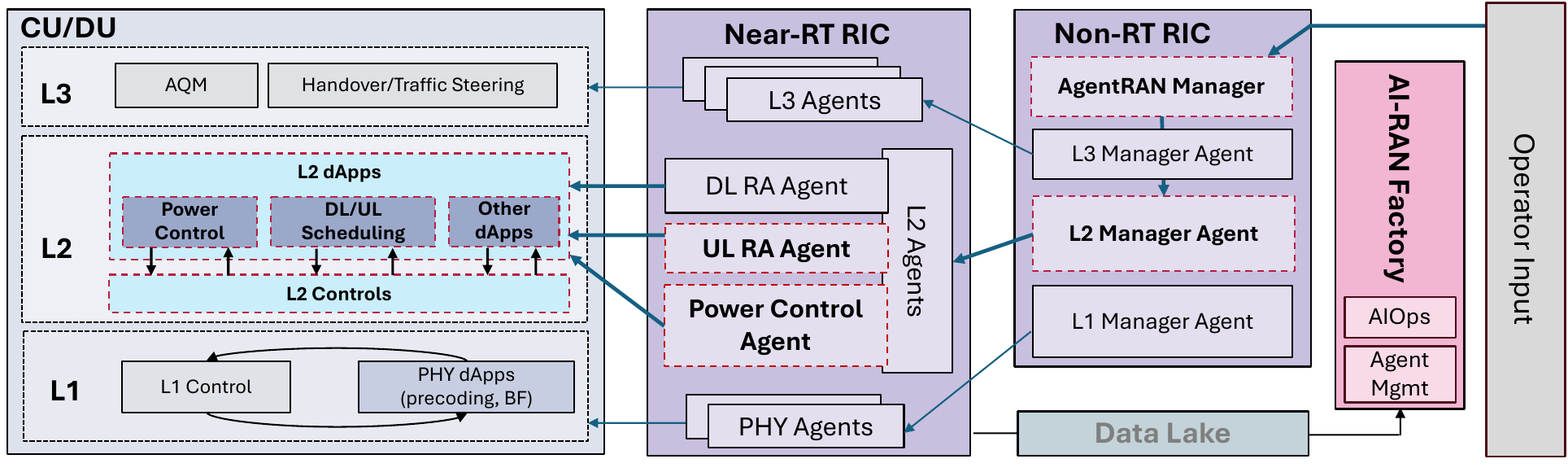}
    \caption{\framework high-level architecture.}
    \label{fig:architecture}
\end{figure*}

\textbf{The \framework Vision.} \rev{O-RAN introduces the concept of third-party applications that can monitor and control the \gls{ran}: \glspl{rapp} for policy-level decisions ($>1$\:s), \glspl{xapp} for near-real-time control ($10$\:ms--$1$\:s), and \glspl{dapp} for real-time operations ($<10$\:ms).} We propose \framework, a novel paradigm rooted in recent advances in generative \gls{ai} that 
\removed{defines the vision and design of a new cellular architecture enabling} 
\rev{enables} \textbf{hierarchical intent decomposition} 
\removed{across three critical dimensions: (i)~\textbf{timescales}, ranging from non-real-time \glspl{rapp} (policy-level applications, $>1$\:s) to near-real-time \glspl{xapp} ($10$\:ms--$1$\:s) to sub-10\:ms \glspl{dapp} (real-time control logic embedded in the \gls{ran}); (ii)~\textbf{spatial distribution}, coordinating agents across different base stations and network entities; and (iii)~\textbf{protocol stack layers}, translating high-level policies down to physical layer parameters.} 
\rev{across these three time domains, across base stations and network entities (spatial distribution), and across protocol stack layers (from high-level policies to physical layer parameters). Fig.~\ref{fig:architecture} illustrates its architecture.} Our key insight is that while O-RAN specifies \textit{how} agents communicate, \framework specifies \textit{what they communicate and how they act} by introducing standardized \gls{nl}-based coordination protocols. {\bf \framework establishes a semantic and functional complement to O-RAN that builds the foundation for the AI-for-RAN pillar within the broader AI-RAN vision.}

\removed{The \framework architecture, shown in Fig.~\ref{fig:architecture}, shows how high-level operator intents cascade through a hierarchy of AI agents that coordinate via \gls{nl}. Via a recursive parent-child model, each agent leverages \gls{a2a} and \gls{mcp}, both protocols built on JSON-RPC. The former enables agents to expose \glspl{api} to exchange data, including \gls{nl}, files and structured JSON, while the latter is a standardized interface that enables agents to discover and interact with \gls{api} functions through \gls{nl}. \gls{mcp} servers expose network functions (e.g., ``adjust \gls{pc},'' or ``modify scheduling policy'') that agents can discover dynamically and call through standardized \gls{nl} interfaces, enabling flexible integration of new network functions without modifying the agent code. The choice of \gls{nl} as an interface ensures transparency, trust and accountability. Indeed, a human operator can easily read the inter-agent communication, along with the reasoning of each agent, and understand why a decision was made.}

\removed{Figure~\ref{fig:architecture} summarizes a set of possible agents for controlling different network functions.
A subset of agents enclosed in dashed-red lines will serve as a case study later in the article.
In this subset, the L2 manager oversees the \gls{pc} Agent and the \gls{ul} \gls{ra} Agent. These translate high-level intents into specific control parameters executed by their respective \glspl{dapp}, pieces of control logic embedded directly in the \gls{ran} for real-time control. This enables the intent to flow down the network and ultimately influence network controls.}

\removed{For example, the Layer 2 Manager Agent, in the \nonric of Fig.~\ref{fig:architecture} (parent) may try to enforce an intent asking the \gls{ul} and \gls{dl} \gls{ra} agents, in the \nrric (children) to ensure that \gls{urllc} users experience a \gls{rtt} of less than $20$\:ms. 
Although the exact execution of this process might vary from one deployment to another, the following exchange provides an example of such interactions:}

\removed{\begin{dialogue}
\speak{\textcolor{l2color}{L2 Manager Agent (L2)} $\rightarrow$ \textcolor{ulcolor}{\gls{ul} \gls{ra} Agent (\gls{ul} \gls{ra})}}
I need to ensure $20$\:ms of RTT for \gls{urllc}, please report on \gls{ul} performance.

\speak{\textcolor{ulcolor}{UL RA} $\rightarrow$ \textcolor{l2color}{L2}}
There currently is contention with average throughput of $65\:\mathrm{Mbit/s}$. Multiple MTC UEs have poor MCS.

\speak{\textcolor{l2color}{L2} $\rightarrow$ \textcolor{dlcolor}{DL RA Agent (\gls{dl} \gls{ra})}}
How is the resource utilization for DL?

\speak{\textcolor{dlcolor}{DL RA}$\rightarrow$ \textcolor{l2color}{L2}}
Resource usage is low, 43\% average PRB utilization.

\speak{\textcolor{l2color}{L2} $\rightarrow$ \textcolor{dlcolor}{DL RA}}
Serve URLLC UEs with at most $4$\:ms of delay. MTC battery savings are secondary, increase target SNR.

\speak{\textcolor{l2color}{L2} $\rightarrow$ \textcolor{ulcolor}{UL RA}}
Serve URLLC UEs with at most $16$\:ms of delay.

\speak{\textcolor{l2color}{L2} $\rightarrow$ \textcolor{pccolor}{PC Agent (PC)}}
Increase transmission power of MTC and prioritize spectrum efficiency over battery savings.
\speak{\textcolor{ulcolor}{UL RA}, \textcolor{dlcolor}{DL RA}, and \textcolor{pccolor}{PC}}
\textit{Send \glspl{dapp} \gls{mcp} API calls to apply actions.}
\end{dialogue}}

\removed{This case illustrates that the congestion reported by the \gls{ul} agent makes the L2 manager enforce looser delay constraints on the \gls{ul} and stringent constraints on the \gls{dl}.
At the same time, it instructs the \gls{pc} agent to deprioritize the battery saving of \gls{mtc} \glspl{ue} to increase spectral efficiency. Such interactions can be repeated regularly to adapt the policies to the network state.}

\removed{\textbf{Methodology for Hierarchical Context and Agent Coordination.} \framework introduces a systematic approach for managing context across the agent hierarchy. High-level intents are progressively refined as they flow down the protocol stack, with each layer agent maintaining a \textbf{context repository} that aggregates information from its subordinate agents, including current and historical network \glspl{kpi}, ongoing inter-agent negotiations, and constraint propagation rules. Context aggregation rules are dynamically updated based on network performance feedback, enabling agents to learn which information from lower layers is most relevant for decision-making. The framework implements three key coordination mechanisms: (i) \textbf{Intent Cascading}—where high-level operator intents are systematically decomposed into layer-specific sub-intents; (ii) \textbf{Constraint Propagation}—where technical limitations discovered at lower layers inform decisions at higher layers; and (iii) \textbf{Dynamic Policy Negotiation}—where agents can renegotiate \glspl{ra} based on changing network conditions.}

\textbf{Contributions.} Our contributions are as follows.
We define a framework for \gls{ran} agentic control on top of 3GPP and O-RAN systems, with an architecture for on-the-fly agent generation, as well as inter-agent communication and coordination.
We also provide the vision for how the framework can be leveraged for intelligent \gls{ran} bootstrapping, empowering operators with quick \gls{ran} intelligence deployment while improving itself autonomously.
Finally, we demonstrate the feasibility of our agentic \gls{ran} orchestration approach \gls{ota} with a 5G uplink scheduling optimization use case in a real-world private 5G network, \rev{and the AI-RAN factory with an interference mitigation use case.}

\section{From Cloud-Native to \gls{ai}-Native \gls{ran}}

In this section, we first review the recent standardization efforts that enable cloud-native \gls{ran} architectures, providing the foundation for embedding intelligence in cellular networks. We then examine the current landscape of \gls{ai}-based \gls{ran} control, identifying the gaps that \framework addresses.
\subsection{A Primer on \gls{oran} and AI-RAN}
The \gls{oran} ALLIANCE proposes a novel cloud-native cellular architecture. \rev{It introduces the \glsfirst{ric}, a centralized platform that enables intelligent, data-driven control of the \gls{ran}. The \gls{ric} is split into two layers. The \nrric hosts \glspl{xapp}, which are control applications operating at time scales between $10$\:ms and $1$\:s for tasks such as traffic steering and resource optimization. The \nonric hosts \glspl{rapp}, which manage slower, policy-level decisions on time scales longer than $1$\:s~\cite{wani2024open}. These applications implement \textit{control loops} where the application continuously monitors network \glspl{kpi}, makes decisions, applies actions, and observes outcomes. Recently, the \gls{oran} research community is also evaluating real-time (i.e., below $10$\:ms) control opportunities via \glspl{dapp}. These are pieces of control logic embedded directly in the \gls{ran} for sub-millisecond decision-making~\cite{lacava2025dapps}.}
This architecture makes \gls{oran} the ideal platform to enable and embed \gls{ai} within the \gls{ran}, which is the focus of the \gls{ai}-\gls{ran} Alliance. So far, the AI-RAN Alliance has defined three main research axes: (i)~\gls{ai}-and-\gls{ran}, focused on enabling sharing of cloud resources (e.g., \glspl{cpu}, \glspl{gpu}, storage, \gls{ram}) between \gls{ran} and \gls{ai} workloads; (ii)~\gls{ai}-on-\gls{ran} which focuses on \gls{ai} workloads (not necessary related to the \gls{ran}) that run on the spare capacity left by the \gls{ai}-and-\gls{ran} use case; and (iii)~\gls{ai}-for-\gls{ran}, which is the topic most relevant to this article and includes \gls{ai} applications (e.g., r/x/dApps) specifically designed to optimize the \gls{ran}.

\subsection{\gls{ai}-based \gls{ran} control via LLM Agents}

Realizing the \gls{ai}-\gls{ran} vision requires practical control mechanisms that can operate across the r/x/dApp hierarchy. While specialized \gls{ai} models have addressed individual \gls{ran} optimization tasks (e.g., network slicing, and power and handover control),
\glspl{llm} are emerging as uniquely capable of handling the complexity and adaptability required for intent-based network control.
By ingesting large amounts of heterogeneous \gls{nl} data, \glspl{llm} can be
seen as general reasoners that can handle tasks across diverse domains, and can generate responses that are sound depending on the difficulty of each task.
In the \gls{ran} context, most approaches focus on tasks such as querying specifications~\cite{nikbakht2024tspec}, with a few emerging works on \gls{ran} control, mostly limited to simulation scenarios \cite{lotfi2025prompt}.
To our knowledge, the only work that has explored \gls{llm}-based \gls{ran} control with an \gls{ota} testbed is our ALLSTaR framework~\cite{elkael2025allstar}, which focuses on \gls{dl} scheduler optimization.  We also introduced AutoRAN, a system for intent-based network deployment on a real testbed \cite{maxenti2025autoran}.

\rev{Existing AI-RAN approaches fundamentally differ from AgentRAN in that they require predefined parameter interfaces and initial training data before deployment, optimize fixed objectives that require retraining if goals change, and use static architectures. In contrast, AgentRAN accepts \gls{nl} intents, enables immediate deployment of \gls{llm} agents without network-specific training via \gls{icl}, dynamically coordinates control variables across layers and timescales, and generates new agents via the \gls{ai}-\gls{ran} Factory. When specialized models are needed, the Factory can dynamically update them as data drifts. Its hierarchical architecture also allows it to scale as vendors expose additional \gls{ran} controls. This makes direct quantitative comparison difficult as prior work cannot process \gls{nl} intents, which is precisely the capability AgentRAN introduces.}

\section{The \framework Agentic \gls{ran}}



\subsection{Architecture Overview}
\label{sec:arch_overview}
\framework builds on and extends \gls{3gpp} and \gls{oran} specifications to create a coordinated hierarchy of \gls{ai} agents embedded throughout the disaggregated \gls{ran}.
As shown in Fig.~\ref{fig:architecture}, agents operate as r/x/dApps that autonomously close control loops by receiving \glspl{kpi}, processing them through \glspl{llm}, and implementing decisions. \rev{Agents communicate via \gls{a2a} and \gls{mcp}, both built on JSON-RPC. \gls{a2a} enables data exchange (including \gls{nl}, files, and structured JSON), while \gls{mcp} allows agents to dynamically discover and invoke network functions (e.g., ``adjust power control'' or ``modify scheduling policy'') through \gls{nl} interfaces. This coordination follows a systematic protocol: agents first discover available actions via \gls{mcp}, then exchange intents and status reports via \gls{a2a}, and finally propagate control actions to their \glspl{dapp}.}

\rev{Fig.~\ref{fig:architecture} highlights in red the subset of agents used in our experimental validation: an L2 Manager that oversees a \gls{pc} Agent and a \gls{ul} \gls{ra} Agent, each controlling their respective \glspl{dapp}. The following exchange illustrates how these agents could coordinate to meet, e.g., a latency intent:}

\begin{dialogue}
\speak{\textcolor{l2color}{L2 Manager} $\rightarrow$ \textcolor{ulcolor}{\gls{ul} \gls{ra}}}
Ensure $20$\:ms RTT for \gls{urllc}. Report on \gls{ul} performance.

\speak{\textcolor{ulcolor}{UL RA} $\rightarrow$ \textcolor{l2color}{L2}}
Contention present, $65\:\mathrm{Mbit/s}$ average. Multiple MTC UEs have poor MCS.

\speak{\textcolor{l2color}{L2} $\rightarrow$ \textcolor{dlcolor}{DL RA}}
Report DL resource utilization.

\speak{\textcolor{dlcolor}{DL RA} $\rightarrow$ \textcolor{l2color}{L2}}
Low usage, 43\% average PRB utilization.

\speak{\textcolor{l2color}{L2} $\rightarrow$ \textcolor{dlcolor}{DL RA}, \textcolor{ulcolor}{UL RA}, \textcolor{pccolor}{PC}}
DL: serve URLLC with $\leq 4$\:ms delay. UL: $\leq 16$\:ms. PC: increase MTC power, prioritize spectral efficiency.

\speak{\textcolor{ulcolor}{UL RA}, \textcolor{dlcolor}{DL RA}, \textcolor{pccolor}{PC}}
\textit{Execute via \gls{mcp} calls to \glspl{dapp}.}
\end{dialogue}

\rev{This example shows the L2 Manager gathering status from subordinate agents, then distributing sub-intents based on observed conditions, allocating the latency budget to favor DL (which has spare capacity) while instructing the PC agent to optimize spectral efficiency. The use of \gls{nl} ensures that operators can audit any stage of the decision process.}

\subsection{Core Design Principles}
\framework translates high-level operator intents into network actions through hierarchical decomposition across three dimensions, each mapped to \gls{oran} components:

\textbf{Timescale Decomposition.} We leverage \gls{oran} timescale separation: \glspl{rapp} (non-real-time) manage policies, e.g., \gls{ra} priorities; \glspl{xapp} (near-real-time) handle adaptations like traffic steering and set targets, e.g., throughput target; and \glspl{dapp} (real-time) execute operational decisions. 

\textbf{Geographic Distribution.} \glspl{rapp} are centralized in the network-wide \nonric, while \glspl{xapp} are distributed across Near-Real-Time \glspl{ric}, each managing a cluster of \glspl{gnb}, and \glspl{dapp} are directly embedded in each \gls{cu}/\gls{du}.
This enables intent propagation from network-wide optimization down to cell-specific implementations while maintaining local autonomy and responsiveness.

\textbf{Protocol Stack Layering.} We follow the protocol stack layering, with layer agents (L1/L2/L3) managing their respective functionalities and the \framework manager coordinating cross-layer optimization. Each layer agent decomposes network-layer intents into protocol-specific sub-intents, ensuring that high-level objectives like ``minimize latency for \gls{urllc}'' are broken down into multiple actionable controls.

\subsection{Agent Coordination Mechanisms}
\rev{The effectiveness of \framework depends on three coordination mechanisms that enable hierarchical intent management:}

\removed{The effectiveness of \framework depends on the coordination protocol illustrated in Fig.~\ref{fig:diagram}. When an \gls{xapp} agent is added to the network, it discovers the actions it can send to the \glspl{dapp} it is connected to via \gls{mcp}. Then, the \gls{xapp} initiates loops where it regularly queries fresh \glspl{kpi} from the \glspl{dapp}. 
This enables the \gls{xapp} to receive sub-intents from its layer \gls{rapp} and to respond with a context report, which the \gls{rapp} uses to issue refined sub-intents (see dialog example in the introduction of this paper). The \gls{xapp} then starts a loop where it uses the stored \glspl{kpi} to take the right control actions.

This protocol enables (i) \textit{Context Aggregation} through the rolling \gls{kpi} history that flows upward as high-level context reports; (ii) \textit{Intent Cascading}, where abstract operator goals decompose into concrete control actions; and (iii) \textit{Constraint Propagation}, where discovered limitations at lower levels inform feasibility assessments at the \glspl{rapp}, which refine their intents and intent breakdown logic. 
The use of \gls{nl}
maintains transparency—operators can audit any stage of the decision process—while \gls{mcp}'s standardized interface allows new network functions to be integrated without modifying agent code.
}

\rev{\textbf{Context Aggregation.} Each agent maintains a rolling history of \glspl{kpi} from its subordinate agents, which flows upward as high-level context reports. This enables higher-layer agents to make informed decisions based on ground-truth conditions observed at lower layers.}

\rev{\textbf{Intent Cascading.} High-level operator intents are systematically decomposed as they flow down the hierarchy. Each agent interprets intents within its operational domain and generates appropriate sub-intents for its subordinates, as illustrated in the dialogue example in Section~\ref{sec:arch_overview}.}

\rev{\textbf{Constraint Propagation.} Technical limitations discovered at lower layers (e.g., channel conditions, hardware capabilities) inform feasibility assessments at higher layers. This feedback allows \glspl{rapp} to refine intents based on capabilities, and enables \textit{intent assurance} to detect when the operational state deviates from the target state (known as \textit{intent drift}) and taking corrective actions to restore alignment~\cite{dzeparoska2024intent}.
}
\begin{table*}[h]
\scriptsize
\centering
\setlength{\tabcolsep}{4.5pt}
\setlength\abovecaptionskip{2pt}
\begin{tabular}{p{0.31\textwidth}|p{0.35\textwidth}|p{0.29\textwidth}}
\toprule
\textbf{Original Intent} & \textbf{Emergency Intent} & \textbf{Post-Emergency Intent}\\
\midrule
\textbf{Layer 2 Manager:} Maximize the overall throughput of the system and do not throttle any user or try to save battery. & \textbf{Layer 2 Manager:} There was a life-emergency: all MTC sensors are high priority and need $30\:\mathrm{Mbit/s}$ minimum. & \textbf{Layer 2 Manager:} Incident finished. FWA should be prioritized with high spectral efficiency. MTC has $5\:\mathrm{Mbit/s}$ 
and needs to save lots of battery. \\
\textbf{Power Agent:} Set high target SNR for both MTC and FWA classes to maximize throughput without considering battery consumption. & \textbf{Power Agent:} Due to life-emergency: Increase target SNR for MTC sensors to ensure reliable communication and meet the $30\:\mathrm{Mbit/s}$ minimum requirement per MTC device. & \textbf{Power Agent:} FWA should have high power for high spectral efficiency. MTC should have low power to save battery. \\
\textbf{Scheduler Agent:} Remove throughput limits for both MTC and FWA classes to allow maximum possible throughput while maintaining QoS requirements. & \textbf{Scheduler Agent:} Due to life-emergency: Set MTC sensors as high priority and allocate sufficient resources to guarantee minimum $30\:\mathrm{Mbit/s}$ throughput per MTC device. & \textbf{Scheduler Agent:} FWA should be prioritized for scheduling. MTC requires $5\:\mathrm{Mbit/s}$ throughput. \\
\bottomrule
\end{tabular}
\caption{Network intent evolution.}
\label{tab:intents}
\vspace{-10pt}
\end{table*}
\subsection{Self-Learning Networks: The AI-RAN Factory}

Another major feature of \framework is the integration of the data lake and the introduction of the \gls{ai}-\gls{ran} factory. The data lake provides a persistent repository for all network \glspl{kpi} and agents' decisions, enabling scalable storage and annotation of operational data.

\textbf{Continuous Intelligence Evolution.} The \gls{ai}-\gls{ran} factory represents a paradigm shift toward networks that continuously improve their intelligence. By consuming data from the data lake, it performs analytics to identify improvement opportunities such as sub-optimal agent decisions, configuration errors, unresolved conflicts, or resource-intensive agents. This analysis drives the automatic generation of enhanced agents (e.g., deploying a faster model for time-critical applications).

The factory's analysis pipeline processes both structured data (KPIs, control decisions, performance metrics) and unstructured data (A2A agent conversations, reasoning traces) to identify optimization opportunities. This rich dataset enables the factory to understand not just what decisions were made, but why agents made them and how they negotiated trade-offs. To address this diversity of data sources, the factory integrates a mix of classical statistical analysis tools with an AI agent capable of understanding unstructured data. This agent synthesizes insights, produces  reports, and triggers actions to improve network operations, for example by training and deploying new agents in real time. In this way, the AI-RAN Factory transforms the network into a self-learning system, evolving its intelligence over time.

\textbf{Dynamic Agent Generation.} The \gls{ai}-\gls{ran} factory generates new agents by leveraging the following technologies:
\begin{itemize}
    \item \textit{Code generation}—as demonstrated in ALLSTaR~\cite{elkael2025allstar}, the factory can analyze data collected both \gls{ota} from deployed agents and \glspl{dapp}, as well as from literature papers, identify successful algorithms, and generate entirely new \gls{dapp} implementations combining the best aspects of different algorithms while adhering to operator intents. For example, ALLSTaR processes data from \gls{ota} tests of different schedulers, identifies the best-performing ones, and combines them to create new intent-based schedulers.
    \item \textit{Model distillation}—when \gls{llm} agents consistently make similar decisions for specific scenarios, the factory detects it through its analytics module and distills this knowledge into smaller, faster models while maintaining decision quality.
    \item \textit{Fine-tuning}—when a model does not reach satisfactory performance, the factory initiates fine-tuning of the model. Compared to distillation, fine-tuning does not imitates existing models, but instead adds new training data to update the weights of the model and improve its effectiveness, e.g., via offline \gls{drl} training~\cite{tsampazi2024pandora}.
    \item \textit{Hybrid agent creation}—by combining the above techniques to obtain agents that are both lightweight and effective. For example, a \gls{drl} model can be a subroutine of an \gls{llm} agent to perform its tasks under known conditions, while falling back to the fully \gls{llm}-based approach when there is a new intent or the operational conditions are unseen.
\end{itemize}

These generation strategies work in concert with the factory's continuous monitoring capabilities. For instance, \rev{if the operator's intent specifies that spectral efficiency takes precedence during peak hours,} the analytics module might detect improper behavior where the PC agent consistently reduces MTC power by 3 dB during \removed{peak hours. This is a problem as, at peak hours, spectral efficiency is more important than energy savings.}\rev{those periods, contradicting the stated priority.}
Leveraging this insight, the factory initiates fine-tuning to teach the agent to prioritize spectral efficiency during high-load periods while still preserving battery during off-peak hours. 
After applying the appropriate generation strategy—whether code generation, distillation, fine-tuning, or hybrid creation—the factory validates the resulting agent in a sandbox environment to ensure it meets performance requirements before deployment.

\textbf{Autonomous Network Bootstrapping.} This architecture dramatically reduces the complexity of deploying \gls{ran} optimization algorithms. Unlike traditional \gls{ai} approaches that require extensive data collection campaigns before deployment, operators can deploy \gls{llm} agents from day one, leveraging their general reasoning capabilities without the need for network-specific training data, \rev{i.e., thanks to \gls{icl} \cite{dong2024survey}}. \rev{\gls{icl} also grants them a degree of robustness to out-of-distribution network situations (for example, as will be seen in our experimental section, the \gls{llm} is able to adapt its controls to scenarios it has not been specifically trained for). This has the potential to adress a} major challenge in traditional specialized \gls{ai} models that often fail when encountering unseen network conditions. As the network operates, the \gls{ai}-\gls{ran} factory enables continuous optimization through data-driven enhancements, creating a seamless pathway to optimize performance from immediate deployment. This bootstrapping approach also facilitates the hybrid integration strategy described above, where specialized \gls{ai} models can be gradually integrated under LLM coordination as they become available.

For example, an operator could deploy an large \gls{llm}-based agent (e.g., Qwen3-32B), which provides satisfactory policies but consumes large computational resources. The \gls{ai}-\gls{ran} factory can then distill this knowledge into a smaller model (e.g., Qwen3-4B) or traditional \gls{ai} algorithms using imitation learning. Beyond model distillation, the factory can also generate entirely new network functions through code generation and algorithmic recombination. We have demonstrated this 
factory cycle in ALLSTaR~\cite{elkael2025allstar}, where LLM-generated schedulers from the academic literature are automatically tested \gls{ota}, and successful control algorithms are recombined into specialized \gls{ibn} schedulers that achieve both high performance and intent compliance when deployed.

\section{\removed{Over-the-Air Agentic Uplink Use Case }\rev{Experimental Demonstration}}
\label{sec:usecase}

We now demonstrate \framework's feasibility and effectiveness through a real-world, O-RAN-compliant deployment. \removed{The following section presents an \gls{ota} prototype that showcases how hierarchical intent decomposition enables dynamic uplink optimization across power control and scheduling domains, thereby validating \framework's  capability to translate high-level operator intents into coordinated, cross-domain network actions.}
\rev{We present two experiments: (i) an over-the-air deployment demonstrating intent-driven multi-agent coordination for uplink optimization; and (ii) a channel emulator setup validating AI-RAN Factory self-improvement under distribution shift in an interference detection setup. All agents use Claude Sonnet 4 via API with no network-specific fine-tuning; decision logic emerges from prompts containing each agent's role, current intent, recent \glspl{kpi}, action bounds, and decision history.}
\removed{
\subsection{Problem Description and Considered Architecture}
Scheduling the 5G \gls{ul} involves multiple decision variables (transmission power, \glspl{prb} allocation, \gls{mcs}) that must adapt to changing channel and traffic conditions, creating tradeoffs between energy-efficiency, throughput, QoS, fairness, and delay. Since 3GPP leaves scheduling algorithms to manufacturers' discretion, \gls{ul} scheduling is ideal for demonstrating \framework. To illustrate how \framework manages these competing objectives and decision variables, we present our solution approach following the red portions of Fig.~\ref{fig:architecture} from right to left, beginning with the \framework manager.

In this use-case, since we consider one layer agent only (the L2 Manager Agent), the \framework agent directly passes the intent to the L2 agent. Then, the L2 agent takes that intent and breaks it down into two sub-intents, one for the \gls{pc} agent and one for the \gls{ul} \gls{ra} agent. 
The \gls{pc} agent then parses the intent and periodically (e.g., every second) queries the network for \glspl{kpi}. Based on these \glspl{kpi}, it decides which \gls{pc} target to set for each \gls{ue} (i.e., the target \gls{snr} level). This target is passed to the \gls{pc} \gls{dapp}, which, at every \gls{ul} slot, evaluates the current \gls{snr} and decides whether to send \gls{tpc} commands to the \gls{ue} asking it to change its transmit power by $+1$, $-1$, $0$, or $+3$\:dB.
In parallel, the second intent from the L2 manager is sent to the \gls{ul} \gls{ra} agent. From this intent and the periodic \glspl{kpi}, it selects a new scheduling policy which is sent to the \gls{dl}/\gls{ul} \gls{ra} \gls{dapp}, which is tasked to deploy the new \gls{ul} scheduling logic inside the \gls{ul} scheduling \gls{dapp}. 
In our prototype, the \gls{ul} agent updates a modified proportional fair scheduler in which different maximum throughput values per \gls{ue} (thereafter called throttling limits)  can be defined for a whole slice (i.e., class of users). Note that more complex algorithms, such as \gls{ibn} schedulers of~\cite{elkael2025allstar} could be used in this \gls{dapp} out-of-the-box.
}

\begin{figure*}[htbp]
\vspace{-10pt}
    \centering
    \begin{subfigure}[b]{0.32\textwidth}
    \setlength\abovecaptionskip{-7.4pt}
        \centering
            \setlength\fwidth{0.9\linewidth}
            \setlength\fheight{4cm}
            \input{figures/fwa_performance}
        \caption{FWA \glspl{ue}}
        \label{fig:fwa}
    \end{subfigure}
    \hspace{12pt}
    \begin{subfigure}[b]{0.32\textwidth}
    \setlength\abovecaptionskip{-7.4pt}
        \centering
            \setlength\fwidth{0.9\linewidth}
            \setlength\fheight{4cm}
            \input{figures/mtc_performance}
        \caption{MTC \gls{ue}}
        \label{fig:mtc}
    \end{subfigure}
    \hfill
    \begin{subfigure}[b]{0.32\textwidth}
        \centering
            \setlength\fwidth{0.9\linewidth}
            \setlength\fheight{4cm}
            \input{figures/power_consumption}
        \caption{MTC \gls{ue}}
        \label{fig:power}
    \end{subfigure}
    \caption{Impact of agent decisions on UEs during normal phase ($< 300$\:s), emergency ($300$-$600$\:s), and post-emergency ($> 600$\:s).}
    \label{fig:combined}
    \vspace{-5pt}
\end{figure*}
\removed{\subsection{Scenario}}
\removed{
We consider a scenario comprised of two slices. The first slice is \gls{fwa} and has two \glspl{ue}. These \glspl{ue} are interested in high throughput, while power consumption is a minor concern as they are directly connected to the electrical grid.
The second class is \gls{mtc} battery-powered CCTV cameras that operate at low throughput but may require high-quality video streaming during emergencies, inducing higher energy consumption compensated later via energy harvesting (e.g., solar panels).
At the beginning, devices operate under normal conditions and the network uses a proportional fairness scheduling algorithm. Then, because of an emergency, the operator prompts \framework to prioritize \gls{mtc} users. Finally, after the emergency, the operator asks \framework to minimize the energy consumption of \gls{mtc} \glspl{ue} as high video quality is not needed anymore. Note that the intents and sub-intent are fed to each agent with a larger system prompts which contains information such as their role, the context, the bounds of the output they can give (e.g., we limit the \gls{pc} agent to changes of at most $\pm 3$\:dB per step) and a history of the recent decisions the agent has taken, along with the resulting network \glspl{kpi}.

\subsection{Experimental Setup}
We implement \framework for \gls{ul} control as an extended version of \gls{oai} that contains the \gls{dl} scheduling \gls{dapp} from our previous ALLSTaR work~\cite{elkael2025allstar}. 
We significantly refactor the \gls{ul} scheduler of \gls{oai} to leverage logic written in Lua programming language on a dynamically loaded script, similar to the \gls{dl} of ALLSTaR. We also extend \gls{oai} to enable our \gls{pc} \gls{dapp}.
In vanilla \gls{oai}, the \gls{snr} target is a parameter contained in a configuration file and loaded at cell start. Instead, we extend \gls{oai} so this parameter can be changed dynamically and per-\gls{ue}. 
Our \gls{ai} agents use Claude 4 Sonnet API calls with \glspl{kpi} and controls exchanged through an E2/A1-like Redis interface. Agents receive 10-sample sliding windows of historical \glspl{kpi} with guardrails including ±3dB per-cycle \gls{snr} limits (absolute bounds [-15, 18]dB) and throttling limits between 3-100:Mbit/s. \rev{Each agent's prompt contains its role, the current intent, recent \glspl{kpi}, action bounds, and its recent decision history; the agent outputs the next control action (e.g., target \gls{snr} or throttling limit) based on this context. The L2 agent additionally receives the list of available sub-agents and their control capabilities to perform intent decomposition. No network-specific training is performed—decision logic emerges from the \gls{llm}'s general reasoning capabilities.}

We run \framework using Nvidia-accelerated Aerial Layer 1 on our X5G testbed with an Intel Xeon 6240R server, Nvidia A100 GPU, and Foxconn RPQN RU configured for 2x2 MIMO at 100\:MHz in band N78.
In our setup, we use 3~commercial Sierra Wireless EM9293 \glspl{ue} in an indoor laboratory setup. Finally, since the use case involves energy efficiency of the \gls{mtc} \gls{ue}, we monitor the energy consumption of the \gls{ue} using a YoctoWatt board connected to a mini PC. The YoctoWatt sits between the wall power socket and the AC/DC converter used to power the \gls{ue}, and it enables us to collect energy measurements at a $0.1$\:ms granularity.

\subsection{Experimental Results}
In Tab.~\ref{tab:intents}, we show the set of sub-intents derived by the L2 manager from the original intents fed by the operator. 
We can observe that the L2 manager consistently derives correct sub-intent for each intent, i.e., for each case, the necessary information (such as the \gls{qos} requirements to meet and the slice to prioritize) is passed on to the sub-agent. For example, in the first two intents, it correctly sets the transmission power to the maximum, and reduces it when battery savings are mentioned in the third intent. In that same third intent, high spectral efficiency is correctly interpreted for \gls{fwa} as the need for higher target \gls{snr}.
During our experimentation, we tried running the breakdown process of the L2 Manager 20 times and we observed that all of the sub-intents derived during those runs were consistent (contained the information necessary to satisfy the intent).

In Fig.~\ref{fig:combined}, we showcase the decisions made by \framework over time for each slice, as well as their impact on \gls{ue} \glspl{kpi}. We see that at first, when the original intent is used (until second $300$), \framework does not try to limit either the throughput or the energy consumption of the \glspl{ue}, resulting in fair allocation with all \glspl{ue} achieving $20\:\mathrm{Mbit/s}$ of \gls{ul} throughput. Then, in the emergency phase (seconds $300$ to $600$), the \gls{fwa} \glspl{ue} are throttled via the scheduler agent (Fig.~\ref{fig:fwa}). 
This has the effect of limiting the amount of traffic they send, which leaves more capacity (up to $30-40\:\mathrm{Mbit/s}$, which satisfies the intent requirement) for the \gls{mtc} slice (Fig.~\ref{fig:mtc}).

In the meantime, the throughput throttling limit of \gls{fwa} is quickly increased, enabling the \glspl{ue} belonging to this slice to reach their pre-emergency throughput.
We observe adaptive behavior around 350 s, where the scheduler agent gradually increases the \gls{fwa} throttling limit, trying to squeeze more performances while staying within the constraints imposed by \gls{mtc}. This first causes improvements of \gls{fwa} while decreasing the \gls{mtc} throughput, which at first remains above the specified minimum. There then is a brief violation of the intent, which the agent detects. It reacts with a stark reduction of the \gls{fwa} throttling limit which restores \gls{mtc} to its required throughput, demonstrating the effectiveness of the \gls{kpi} feedback loop. 

Finally, in the third phase (after second $600$), the agents reduce the target \gls{snr} of \gls{mtc} \glspl{ue}, which results in them saving around $200$\:mW of power compared to the emergency phase (Fig.~\ref{fig:power}).
During the final intent transition at 600 s, agents hit the $\pm3$ dB \gls{snr} adjustment guardrails, maintaining maximum allowed changes for multiple cycles to implement a rate-limited trajectory toward their targets.
Overall, these results show that \framework is capable of swiftly adapting to new situations solely based on high-level \gls{nl} intents.
}
\rev{
\subsection{Experiment 1: Intent-Driven Uplink Coordination}
}
\textbf{Problem and Architecture.} Scheduling the 5G \gls{ul} involves multiple decision variables (transmission power, \glspl{prb} allocation, \gls{mcs}) that must adapt to changing channel and traffic conditions, creating tradeoffs between energy-efficiency, throughput, QoS, fairness, and delay. Since 3GPP leaves scheduling algorithms to manufacturers' discretion, \gls{ul} scheduling is ideal for demonstrating \framework. 
To illustrate how \framework manages these competing objectives, we describe our solution following the red portions of Fig.~\ref{fig:architecture} from right to left. 
In this use case, the \framework manager passes the intent directly to the L2 agent, which decomposes it into two sub-intents: one for the \gls{pc} agent and one for the \gls{ul} \gls{ra} agent. The \gls{pc} agent periodically (e.g., every second) queries network \glspl{kpi} and decides which target \gls{snr} level to set for each \gls{ue}. This target is passed to the \gls{pc} \gls{dapp}, which evaluates the current \gls{snr} at every \gls{ul} slot and issues \gls{tpc} commands to adjust \gls{ue} transmit power by $+1$, $-1$, $0$, or $+3$\:dB. In parallel, the \gls{ul} \gls{ra} agent selects scheduling policies deployed via a \gls{ra} \gls{dapp}. In our prototype, this \gls{dapp} updates a modified proportional fair scheduler in which different maximum throughput values (throttling limits) can be defined per slice.

\textbf{Scenario.} We consider two slices. The first is \gls{fwa} with two \glspl{ue} interested in high throughput; power consumption is a minor concern as they are grid-powered. The second is \gls{mtc} battery-powered CCTV cameras that normally operate at low throughput but may require high-quality video streaming during emergencies, inducing higher energy consumption later compensated via energy harvesting.

The experiment proceeds in three phases. Initially, devices operate under normal conditions with proportional fairness scheduling. At 300s, an emergency occurs and the operator prompts \framework to prioritize \gls{mtc} users. Finally, at 600 s, the operator asks \framework to minimize \gls{mtc} energy consumption since high-quality video is no longer needed. Note that intents and sub-intents are provided to each agent within a larger system prompt containing their role, context, output bounds (e.g., the \gls{pc} agent is limited to $\pm 3$\:dB changes per step), and a history of recent decisions with resulting \glspl{kpi}.

\textbf{Setup.} We implement \framework as an extended version of \gls{oai} incorporating the \gls{dl} scheduling \gls{dapp} from ALLSTaR~\cite{elkael2025allstar}. We significantly refactor the \gls{ul} scheduler to leverage dynamically-loaded Lua scripts, similar to ALLSTaR's \gls{dl}. We also extend \gls{oai} to enable dynamic per-\gls{ue} \gls{snr} targets, replacing the static parameter.

Our agents use Claude Sonnet 4 API calls with \glspl{kpi} and controls exchanged through a Redis interface. Agents receive 10-sample sliding windows of historical \glspl{kpi} with guardrails including $\pm3$dB per-cycle \gls{snr} limits (absolute bounds $[-15, 18]$dB) and throttling limits between 3--100\:Mbit/s. \rev{Each agent's prompt contains its role, current intent, recent \glspl{kpi}, action bounds, and decision history; the agent outputs the next control action based on this context. The L2 agent additionally receives the list of available sub-agents and their control capabilities to perform intent decomposition. No network-specific training is performed—decision logic emerges from the \gls{llm}'s general reasoning capabilities.}

We deploy on our X5G testbed using Nvidia Aerial Layer~1 with an Intel Xeon 6240R server, Nvidia A100 GPU, and Foxconn RPQN RU configured for 2x2 MIMO at 100\:MHz in band N78. Three commercial Sierra Wireless EM9293 \glspl{ue} operate in an indoor laboratory setup. We monitor \gls{mtc} energy consumption using a YoctoWatt board at 0.1\:ms granularity, connected between the wall socket and the \gls{ue}'s AC/DC converter. \rev{While our results illustrate only one experimental run, we validate consistency of both \glspl{kpi} and sub-intent breakdown over 10 different experimental runs.}

\textbf{Results.} Tab.~\ref{tab:intents} shows the sub-intents derived by the L2 manager from operator intents. The L2 manager consistently extracts correct sub-intents: necessary information such as QoS requirements and slice priorities is properly passed to sub-agents. For example, transmission power is set to maximum in the first two intents but reduced when battery savings are mentioned in the third. High spectral efficiency is correctly interpreted for \gls{fwa} as requiring higher target \gls{snr}. Across 10 trials, all derived sub-intents contained the information necessary to satisfy the intent.

Fig.~\ref{fig:combined} shows agent decisions' impact on \gls{ue} \glspl{kpi}. In normal operation ($<300$s), no limits are imposed, yielding fair 20\:Mbit/s allocation. During emergency (300--600 s), the scheduler agent throttles \gls{fwa} \glspl{ue} (Fig.~\ref{fig:fwa}), freeing capacity for \gls{mtc} to reach 30--40\:Mbit/s (Fig.~\ref{fig:mtc}), satisfying the intent.

We observe adaptive behavior around 350 s: the scheduler agent gradually increases the \gls{fwa} throttling limit, attempting to improve \gls{fwa} performance while respecting \gls{mtc} constraints. This initially succeeds, but briefly causes \gls{mtc} throughput to drop below requirements. The agent detects this violation through \gls{kpi} feedback and reacts with a sharp reduction of the \gls{fwa} throttling limit, restoring \gls{mtc} throughput and demonstrating effective closed-loop control.

In the third phase ($>600$ s), agents reduce \gls{mtc} target \gls{snr}, saving approximately 200\:mW compared to the emergency phase (Fig.~\ref{fig:power}). During the transition at 600 s, agents hit the $\pm3$ dB guardrails, maintaining maximum allowed changes for multiple cycles to implement a rate-limited trajectory toward their targets. Overall, these results demonstrate that \framework swiftly adapts to changing conditions based solely on high-level \gls{nl} intents.

\rev{
\subsection{Experiment 2: AI-RAN Factory Self-Improvement}


\textbf{Scenario.}~To validate the AI-RAN Factory's autonomous retraining under distribution shifts, we consider interference management in the shared-spectrum deployment (e.g., CBRS) in Fig. \ref{fig:scenario}, where transmissions from \gls{gnb} A may cause harmful interference to neighboring \glspl{gnb} belonging to another operator (gNB D and E) on overlapping CBRS channels (e.g., 5\:MHz of bandwidth). Two additional neighboring cells (gNB B and C) belong to the same operator as gNB A. The system receives only binary alerts from the other operator, sent when interference exceeds -70\:dB at one of its \glspl{gnb}, indicating that interference must be reduced. To address this, an xApp is trained to predict when transmissions will trigger an alert, based on the RSRP of \gls{gnb} A and the \gls{ul} received power at \gls{gnb} B and C. When an alert is predicted, a coordinated xApp preemptively blocks the overlapping PRBs for the affected \gls{ue}, mitigating interference before complaints arise. The model is initially trained on data collected with a with a \gls{ue} moving north to south (looping over points 1-2-3). A distribution drift is then introduced by having the \gls{ue} move west, triggering prediction errors that must be caught by the AI-RAN Factory, which should initiate retraining.

\begin{figure}[htbp]
    \centering
    \resizebox{\columnwidth}{!}{\input{figures/result2}}
    \caption{Propsim-based interference prediction scenario.}
    \label{fig:scenario}
\end{figure}

\textbf{Setup.} We use the same \gls{gnb} and \gls{ue} as Experiment~1, but replace the \gls{ota} channel with a Propsim channel emulator configured with a free-space path loss model to enable repeatable mobility experiments. \glspl{gnb} B-E are emulated using four channels of a USRP X410 that measure received signal strength. Only \gls{gnb} A, B and C measurements are exposed to \framework, along with the alerts.
We first train a decision tree which predicts every 100\:ms whether an alert will happen in the next second (using the north-south data i.e. the UE moving between points 1-2-3 in a loop), before deploying it as an xApp with performance monitored by the L2 manager.

During operation, the factory LLM continuously monitors prediction accuracy by comparing xApp outputs against actual alerts. If an alert is predicted, the xApp triggers the resource allocation dApp to block the 5\:MHz of overlapping \glspl{prb}. We repeat the experiment 10 times for validation and report average accuracies thereafter. The LLM also has the option to trigger the retraining of the offloading model at any time.

\textbf{Results.} While on the pre-training path, the xApp achieves 97\% prediction accuracy (as indicated by color shades of UE markers in Fig. \ref{fig:scenario}), effectively preventing most alerts through proactive mitigation.
When \gls{ue} mobility switches to the unseen area (point 3-4 to 7), i.e., shifting distribution, accuracy drops significantly to 43\% between points 4 and 5. This means we fail to anticipate alert and cause interference. The AI-RAN Factory's LLM agent detects this degradation through increased alerts despite active mitigation. It autonomously initiates the retraining loop: the AI-RAN factory leverages fresh data collected in the data lake (associating RSRP measurements with alerts), regenerates the prediction model incorporating both scenarios, validates that accuracy improves significantly, and deploys the updated xApp. Post-retraining, the system achieves 95\% prediction accuracy on both scenarios, demonstrating successful adaptation to new conditions without operator intervention (as seen with the high detection rate in position 6, 9 and 10).
This validates the AI-RAN Factory's ability to automatically detect out-of-distribution conditions and retrain agents, enabling self-improvement as network conditions evolve.
}

\section{Emergent Behaviors in \gls{nl}-Based Control}

Our experiments reveal that \gls{llm}-based network control exhibits emergent behaviors not explicitly programmed into the system. When agents encounter operational constraints like the $\pm 3$\:dB adjustment limit, they autonomously develop multi-cycle probing strategies to reach their targets as close as possible to the intent. This suggests that \glspl{llm} can discover appropriate control strategies through interaction rather than requiring explicit algorithmic specification.
The ability of agents to understand sensitive controls, such as the throughput-power tradeoff for MTC devices, through contextual observation raises questions about the nature of network control. Traditional approaches require either complex modeling or extensive parameter tuning to capture these relationships. In contrast, our agents adapt this understanding through observation of measured outcomes alone.
However, this flexibility comes with trade-offs. The non-deterministic nature of LLM decisions means identical network states might produce slightly different control actions. \rev{We address this through closed-loop \gls{kpi} feedback, which compensates for variability at the outcome level, and guardrails that bound the action space. For safety-critical functions requiring highly predictable behavior, the \gls{ai}-\gls{ran} Factory could distill consistent decisions into near-deterministic or bayesian models.} \removed{While our results show that closed-loop feedback compensates for this variability at the outcome level, further} \rev{Further} investigation is needed to understand the implications for network stability guarantees and coordination between multiple agents. The question remains whether language-based control can meet the stringent requirements of production networks, particularly for safety-critical functions.

\rev{\textbf{Trade-offs of NL Interfaces.} While \gls{nl} coordination is transparent, it introduces trade-offs: intents may be ambiguous, which we mitigate via structured prompts with explicit action bounds. Intents can also be impossible to satisfy, or yield unexpected behavior, which are challenges we leave for future work. \gls{llm} inference also adds latency, acceptable for r/xApps but not for real-time \glspl{dapp}, which is why they take predefined input format and are exposed as \gls{mcp}. Those predefined formats enable operators to analyze control decisions as classical time-series without preventing them from inspecting the \gls{llm}-generated reasoning at the x/rApp.}

\section{Conclusion}
We presented \framework, a novel agentic AI architecture for intent-based cellular networks. With hierarchical intent decomposition across timescales, deployments, and protocol stack layers, \framework enables autonomous coordination through \gls{nl} interfaces.
Our key contributions include: (i) integration of \gls{oran} closed-loop control with agentic \gls{ai}; (ii) the \gls{ai}-\gls{ran} Factory, to bootstrap intelligence after day zero and continuously self-learn; and (iii) \gls{ota} validation.
The experimental results demonstrate rapid adaptation to changing network conditions based solely on \gls{nl} intents, while maintaining transparency and interpretability. Essentially, \framework builds an O-RAN-aligned fabric for the AI-for-RAN pillar within the broader AI-RAN vision, establishing a pathway toward networks that evolve their own intelligence. \rev{Future work includes dynamic context aggregation rules and conflict arbitration mechanisms for multi-agent negotiation.}
\bibliographystyle{IEEEtran}
\bibliography{biblio}






\vspace{-46pt}
\begin{IEEEbiographynophoto}
{Maxime Elkael} is a Postdoctoral Researcher at \gls{neu}. He received his Ph.D. from Institut Polytechnique de Paris in 2023.

\vspace{5pt}
\noindent
\textbf{Salvatore D'Oro} is a Research Associate Professor at \gls{neu} and the CTO at zTouch Networks. He received his Ph.D. from
University of Catania in 2015.

\vspace{5pt}
\noindent
\textbf{Leonardo Bonati} is an Associate Research Scientist at \gls{neu}, where he received his Ph.D. in 2022.

\vspace{5pt}
\noindent
\textbf{Michele Polese} is a Research Assistant Professor at \gls{neu}. He received his Ph.D. from the University of Padova in 2020.

\vspace{5pt}
\noindent
\textbf{Yunseong Lee} conducts R\&D in AI-RAN with \gls{sb}. He received his Ph.D. from Tohoku University in 2020.

\vspace{5pt}
\noindent
\textbf{Koichiro Furueda} conducts research and development in AI-RAN at \gls{sb}.

\vspace{5pt}
\noindent
\textbf{Tommaso Melodia} is the William Lincoln Smith Chair Professor at \gls{neu} and Director of the Institute for Intelligent Networked Systems.
\end{IEEEbiographynophoto}

\end{document}

%% file: acronyms.tex
\newacronym{isac}{ISAC}{Integrated Sensing and Communication}
\newacronym{tpc}{TPC}{Transmit Power Control}
\newacronym{re}{RE}{Resource Element}
\newacronym{ewma}{EWMA}{Exponentially Weighted Moving Average}
\newacronym{gpfb}{GPFB}{Generalized Proportional Fair Buffer}
\newacronym{drb}{DRB}{Data Radio Bearer}
\newacronym{hol}{D-HoL}{Head-of-Line Delay}
\newacronym{bcqi}{BCQI}{Best CQI}
\newacronym{ft}{FT}{Fair Throughput}
\newacronym{rr}{RR}{Round-Robin}
\newacronym{pf}{PF}{Proportional Fair}
\newacronym{x5g}{X5G}{}
\newacronym{ibs}{IBS}{Intent-Based Scheduling}
\newacronym{rapp}{rApp}{RAN Application}
\newacronym{xapp}{xApp}{eXtended Application}
\newacronym{dapp}{dApp}{Distributed Application}
\newacronym{ibn}{IBN}{Intent-Based Networking}
\newacronym{dlsch}{DLSCH}{Downlink Shared Channel}
\newacronym{ocr}{OCR}{Optical Character Recognition}
\newacronym{3gpp}{3GPP}{3rd Generation Partnership Project}
\newacronym{sriov}{SR-IOV}{Single Root I/O Virtualization}
\newacronym{vf}{VF}{Vitual Functions}
\newacronym{4g}{4G}{4th generation}
\newacronym{5g}{5G}{5th generation}
\newacronym{6g}{6G}{6th generation}
\newacronym{5gc}{5GC}{5G Core}
\newacronym{adc}{ADC}{Analog to Digital Converter}
\newacronym{aerpaw}{AERPAW}{Aerial Experimentation and Research Platform for Advanced Wireless}
\newacronym{ai}{AI}{Artificial Intelligence}
\newacronym{aimd}{AIMD}{Additive Increase Multiplicative Decrease}
\newacronym{am}{AM}{Acknowledged Mode}
\newacronym{amc}{AMC}{Adaptive Modulation and Coding}
\newacronym{amf}{AMF}{Access and Mobility Management Function}
\newacronym{aops}{AOPS}{Adaptive Order Prediction Scheduling}
\newacronym{api}{API}{Application Programming Interface}
\newacronym{apn}{APN}{Access Point Name}
\newacronym{ap}{AP}{Application Protocol}
\newacronym{aqm}{AQM}{Active Queue Management}
\newacronym{ausf}{AUSF}{Authentication Server Function}
\newacronym{avc}{AVC}{Advanced Video Coding}
\newacronym{awgn}{AGWN}{Additive White Gaussian Noise}
\newacronym{balia}{BALIA}{Balanced Link Adaptation Algorithm}
\newacronym{bbu}{BBU}{Base Band Unit}
\newacronym{bdp}{BDP}{Bandwidth-Delay Product}
\newacronym{ber}{BER}{Bit Error Rate}
\newacronym{bf}{BF}{Beamforming}
\newacronym{bler}{BLER}{Block Error Rate}
\newacronym{brr}{BRR}{Bayesian Ridge Regressor}
\newacronym{bs}{BS}{Base Station}
\newacronym{bsr}{BSR}{Buffer Status Report}
\newacronym{bss}{BSS}{Business Support System}
\newacronym{ca}{CA}{Carrier Aggregation}
\newacronym{caas}{CaaS}{Connectivity-as-a-Service}
\newacronym{cb}{CB}{Code Block}
\newacronym{cc}{CC}{Congestion Control}
\newacronym{ccid}{CCID}{Congestion Control ID}
\newacronym{cco}{CC}{Carrier Component}
\newacronym{cd}{CD}{Continuous Delivery}
\newacronym{cdd}{CDD}{Cyclic Delay Diversity}
\newacronym{cdf}{CDF}{Cumulative Distribution Function}
\newacronym{cdn}{CDN}{Content Distribution Network}
\newacronym{cli}{CLI}{Command-line Interface}
\newacronym{cn}{CN}{Core Network}
\newacronym{codel}{CoDel}{Controlled Delay Management}
\newacronym{comac}{COMAC}{Converged Multi-Access and Core}
\newacronym{cord}{CORD}{Central Office Re-architected as a Datacenter}
\newacronym{cornet}{CORNET}{COgnitive Radio NETwork}
\newacronym{cosmos}{COSMOS}{Cloud Enhanced Open Software Defined Mobile Wireless Testbed for City-Scale Deployment}
\newacronym{cots}{COTS}{Commercial Off-the-Shelf}
\newacronym{cp}{CP}{Control Plane}
\newacronym{cyp}{CP}{Cyclic Prefix}
\newacronym{up}{UP}{User Plane}
\newacronym{cpu}{CPU}{Central Processing Unit}
\newacronym{cqi}{CQI}{Channel Quality Information}
\newacronym{cr}{CR}{Cognitive Radio}
\newacronym{cran}{CRAN}{Cloud RAN}
\newacronym{crs}{CRS}{Cell Reference Signal}
\newacronym{csi}{CSI}{Channel State Information}
\newacronym{csirs}{CSI-RS}{Channel State Information - Reference Signal}
\newacronym{cu}{CU}{Central Unit}
\newacronym{cubb}{cuBB}{CUDA Baseband}
\newacronym{d2tcp}{D$^2$TCP}{Deadline-aware Data center TCP}
\newacronym{d3}{D$^3$}{Deadline-Driven Delivery}
\newacronym{dac}{DAC}{Digital to Analog Converter}
\newacronym{dag}{DAG}{Directed Acyclic Graph}
\newacronym{das}{DAS}{Distributed Antenna System}
\newacronym{dash}{DASH}{Dynamic Adaptive Streaming over HTTP}
\newacronym{dc}{DC}{Dual Connectivity}
\newacronym{dccp}{DCCP}{Datagram Congestion Control Protocol}
\newacronym{dce}{DCE}{Direct Code Execution}
\newacronym{dci}{DCI}{Downlink Control Information}
\newacronym{dctcp}{DCTCP}{Data Center TCP}
\newacronym{dl}{DL}{Downlink}
\newacronym{dmr}{DMR}{Deadline Miss Ratio}
\newacronym{dmrs}{DMRS}{DeModulation Reference Signal}
\newacronym{drlcc}{DRL-CC}{Deep Reinforcement Learning Congestion Control}
\newacronym{drs}{DRS}{Discovery Reference Signal}
\newacronym{du}{DU}{Distributed Unit}
\newacronym{e2e}{E2E}{end-to-end}
\newacronym{earfcn}{EARFCN}{E-UTRA Absolute Radio Frequency Channel Number}
\newacronym{ecaas}{ECaaS}{Edge-Cloud-as-a-Service}
\newacronym{ecn}{ECN}{Explicit Congestion Notification}
\newacronym{edf}{EDF}{Earliest Deadline First}
\newacronym{embb}{eMBB}{Enhanced Mobile Broadband}
\newacronym{empower}{EMPOWER}{EMpowering transatlantic PlatfOrms for advanced WirEless Research}
\newacronym{enb}{eNB}{evolved Node Base}
\newacronym{endc}{EN-DC}{E-UTRAN-\gls{nr} \gls{dc}}
\newacronym{epc}{EPC}{Evolved Packet Core}
\newacronym{eps}{EPS}{Evolved Packet System}
\newacronym{es}{ES}{Edge Server}
\newacronym{etsi}{ETSI}{European Telecommunications Standards Institute}
\newacronym[firstplural=Estimated Times of Arrival (ETAs)]{eta}{ETA}{Estimated Time of Arrival}
\newacronym{eutran}{E-UTRAN}{Evolved Universal Terrestrial Access Network}
\newacronym{faas}{FaaS}{Function-as-a-Service}
\newacronym{fapi}{FAPI}{Functional Application Platform Interface}
\newacronym{fdd}{FDD}{Frequency Division Duplexing}
\newacronym{fdm}{FDM}{Frequency Division Multiplexing}
\newacronym{fdma}{FDMA}{Frequency Division Multiple Access}
\newacronym{fed4fire}{FED4FIRE+}{Federation 4 Future Internet Research and Experimentation Plus}
\newacronym{fir}{FIR}{Finite Impulse Response}
\newacronym{fit}{FIT}{Future \acrlong{iot}}
\newacronym{fpga}{FPGA}{Field Programmable Gate Array}
\newacronym{fr2}{FR2}{Frequency Range 2}
\newacronym{fs}{FS}{Fast Switching}
\newacronym{fscc}{FSCC}{Flow Sharing Congestion Control}
\newacronym{ftp}{FTP}{File Transfer Protocol}
\newacronym{fw}{FW}{Flow Window}
\newacronym{ge}{GE}{Gaussian Elimination}
\newacronym{gnb}{gNB}{Next Generation Node Base}
\newacronym{gop}{GOP}{Group of Pictures}
\newacronym{gpr}{GPR}{Gaussian Process Regressor}
\newacronym{gpu}{GPU}{Graphics Processing Unit}
\newacronym{gtp}{GTP}{GPRS Tunneling Protocol}
\newacronym{gtpc}{GTP-C}{GPRS Tunnelling Protocol Control Plane}
\newacronym{gtpu}{GTP-U}{GPRS Tunnelling Protocol User Plane}
\newacronym{gtpv2c}{GTPv2-C}{\gls{gtp} v2 - Control}
\newacronym{gw}{GW}{Gateway}
\newacronym{harq}{HARQ}{Hybrid Automatic Repeat reQuest}
\newacronym{hetnet}{HetNet}{Heterogeneous Network}
\newacronym{hh}{HH}{Hard Handover}
\newacronym{hqf}{HQF}{Highest-quality-first}
\newacronym{hss}{HSS}{Home Subscription Server}
\newacronym{http}{HTTP}{HyperText Transfer Protocol}
\newacronym{ia}{IA}{Initial Access}
\newacronym{iab}{IAB}{Integrated Access and Backhaul}
\newacronym{ic}{IC}{Incident Command}
\newacronym{ietf}{IETF}{Internet Engineering Task Force}
\newacronym{imsi}{IMSI}{International Mobile Subscriber Identity}
\newacronym{imt}{IMT}{International Mobile Telecommunication}
\newacronym{iot}{IoT}{Internet of Things}
\newacronym{ip}{IP}{Internet Protocol}
\newacronym{itu}{ITU}{International Telecommunication Union}
\newacronym{kpi}{KPI}{Key Performance Indicator}
\newacronym{kpm}{KPM}{Key Performance Measurement}
\newacronym{kvm}{KVM}{Kernel-based Virtual Machine}
\newacronym{los}{LOS}{Line-of-Sight}
\newacronym{lsm}{LSM}{Link-to-System Mapping}
\newacronym{lstm}{LSTM}{Long Short Term Memory}
\newacronym{lte}{LTE}{Long Term Evolution}
\newacronym{lxc}{LXC}{Linux Container}
\newacronym{m2m}{M2M}{Machine to Machine}
\newacronym{mac}{MAC}{Medium Access Control}
\newacronym{manet}{MANET}{Mobile Ad Hoc Network}
\newacronym{mano}{MANO}{Management and Orchestration}
\newacronym{mc}{MC}{Multi-Connectivity}
\newacronym{mcc}{MCC}{Mobile Cloud Computing}
\newacronym{mchem}{MCHEM}{Massive Channel Emulator}
\newacronym{mcs}{MCS}{Modulation and Coding Scheme}
\newacronym{mec}{MEC}{Multi-access Edge Computing}
\newacronym{mec2}{MEC}{Mobile Edge Cloud}
\newacronym{mfc}{MFC}{Mobile Fog Computing}
\newacronym{mgen}{MGEN}{Multi-Generator}
\newacronym{mi}{MI}{Mutual Information}
\newacronym{mib}{MIB}{Master Information Block}
\newacronym{miesm}{MIESM}{Mutual Information Based Effective SINR}
\newacronym{mimo}{MIMO}{Multiple Input, Multiple Output}
\newacronym{ml}{ML}{Machine Learning}
\newacronym{mlr}{MLR}{Maximum-local-rate}
\newacronym[plural=\gls{mme}s,firstplural=Mobility Management Entities (MMEs)]{mme}{MME}{Mobility Management Entity}
\newacronym{mmtc}{mMTC}{Massive Machine-Type Communications}
\newacronym{mmwave}{mmWave}{millimeter wave}
\newacronym{mpdccp}{MP-DCCP}{Multipath Datagram Congestion Control Protocol}
\newacronym{mptcp}{MPTCP}{Multipath TCP}
\newacronym{mr}{MR}{Maximum Rate}
\newacronym{mrdc}{MR-DC}{Multi \gls{rat} \gls{dc}}
\newacronym{mse}{MSE}{Mean Square Error}
\newacronym{mss}{MSS}{Maximum Segment Size}
\newacronym{mt}{MT}{Mobile Termination}
\newacronym{mtd}{MTD}{Machine-Type Device}
\newacronym{mtu}{MTU}{Maximum Transmission Unit}
\newacronym{mumimo}{MU-MIMO}{Multi-user \gls{mimo}}
\newacronym{mvno}{MVNO}{Mobile Virtual Network Operator}
\newacronym{nalu}{NALU}{Network Abstraction Layer Unit}
\newacronym{nas}{NAS}{Network Attached Storage}
\newacronym{nat}{NAT}{Network Address Translation}
\newacronym{nbiot}{NB-IoT}{Narrow Band IoT}
\newacronym{nfv}{NFV}{Network Function Virtualization}
\newacronym{nfvi}{NFVI}{Network Function Virtualization Infrastructure}
\newacronym{ni}{NI}{Network Interfaces}
\newacronym{llm}{LLM}{Large Language Model}
\newacronym{dpdk}{DPDK}{Data Plane Development Kit}
\newacronym{cicd}{CI/CD}{Continuous Integration and Continuous Delivery/Deployment}
\newacronym{nic}{NIC}{Network Interface Card}
\newacronym{nlos}{NLOS}{Non-Line-of-Sight}
\newacronym{now}{NOW}{Non Overlapping Window}
\newacronym{nsm}{NSM}{Network Service Mesh}
\newacronym[type=hidden]{nr}{NR}{New Radio}
\newacronym[type=hidden]{ota}{OTA}{Over-the-Air}
\newacronym{nrf}{NRF}{Network Repository Function}
\newacronym{nsa}{NSA}{Non Stand Alone}
\newacronym{nse}{NSE}{Network Slicing Engine}
\newacronym{nssf}{NSSF}{Network Slice Selection Function}
\newacronym{o2i}{O2I}{Outdoor to Indoor}
\newacronym{oai}{OAI}{OpenAirInterface}
\newacronym{oaicn}{OAI-CN}{\gls{oai} \acrlong{cn}}
\newacronym{oairan}{OAI-RAN}{\acrlong{oai} \acrlong{ran}}
\newacronym{oam}{OAM}{Operations, Administration and Maintenance}
\newacronym{ofdm}{OFDM}{Orthogonal Frequency Division Multiplexing}
\newacronym{olia}{OLIA}{Opportunistic Linked Increase Algorithm}
\newacronym{omec}{OMEC}{Open Mobile Evolved Core}
\newacronym{onap}{ONAP}{Open Network Automation Platform}
\newacronym{onf}{ONF}{Open Networking Foundation}
\newacronym{onos}{ONOS}{Open Networking Operating System}
\newacronym{oom}{OOM}{\gls{onap} Operations Manager}
\newacronym{opnfv}{OPNFV}{Open Platform for \gls{nfv}}
\newacronym{oran}{O-RAN}{Open RAN}
\newacronym{orbit}{ORBIT}{Open-Access Research Testbed for Next-Generation Wireless Networks}
\newacronym{os}{OS}{Operating System}
\newacronym{oss}{OSS}{Operations Support System}
\newacronym{pa}{PA}{Position-aware}
\newacronym{pase}{PASE}{Prioritization, Arbitration, and Self-adjusting Endpoints}
\newacronym{pawr}{PAWR}{Platforms for Advanced Wireless Research}
\newacronym{pbch}{PBCH}{Physical Broadcast Channel}
\newacronym{pcef}{PCEF}{Policy and Charging Enforcement Function}
\newacronym{pcfich}{PCFICH}{Physical Control Format Indicator Channel}
\newacronym{pcrf}{PCRF}{Policy and Charging Rules Function}
\newacronym{pdcch}{PDCCH}{Physical Downlink Control Channel}
\newacronym{pdcp}{PDCP}{Packet Data Convergence Protocol}
\newacronym{pdf}{PDF}{Probability Density Function}
\newacronym{pdsch}{PDSCH}{Physical Downlink Shared Channel}
\newacronym{pdu}{PDU}{Packet Data Unit}
\newacronym{pgw}{PGW}{Packet Gateway}
\newacronym{phich}{PHICH}{Physical Hybrid ARQ Indicator Channel}
\newacronym{phy}{PHY}{Physical}
\newacronym{pmch}{PMCH}{Physical Multicast Channel}
\newacronym{pmi}{PMI}{Precoding Matrix Indicators}
\newacronym{powder}{POWDER}{Platform for Open Wireless Data-driven Experimental Research}
\newacronym{ppo}{PPO}{Proximal Policy Optimization}
\newacronym{ppp}{PPP}{Poisson Point Process}
\newacronym{prach}{PRACH}{Physical Random Access Channel}
\newacronym{prb}{PRB}{Physical Resource Block}
\newacronym{psnr}{PSNR}{Peak Signal to Noise Ratio}
\newacronym{pss}{PSS}{Primary Synchronization Signal}
\newacronym{pucch}{PUCCH}{Physical Uplink Control Channel}
\newacronym{pusch}{PUSCH}{Physical Uplink Shared Channel}
\newacronym{qam}{QAM}{Quadrature Amplitude Modulation}
\newacronym{qci}{QCI}{\gls{qos} Class Identifier}
\newacronym{qoe}{QoE}{Quality of Experience}
\newacronym{qos}{QoS}{Quality of Service}
\newacronym{quic}{QUIC}{Quick UDP Internet Connections}
\newacronym{rach}{RACH}{Random Access Channel}
\newacronym{ran}{RAN}{Radio Access Network}
\newacronym[firstplural=Radio Access Technologies (RATs)]{rat}{RAT}{Radio Access Technology}
\newacronym{rbg}{RBG}{Resource Block Group}
\newacronym{rcn}{RCN}{Research Coordination Network}
\newacronym{rc}{RC}{RAN Control}
\newacronym{rec}{REC}{Radio Edge Cloud}
\newacronym{red}{RED}{Random Early Detection}
\newacronym{renew}{RENEW}{Reconfigurable Eco-system for Next-generation End-to-end Wireless}
\newacronym{rf}{RF}{Radio Frequency}
\newacronym{rfc}{RFC}{Request for Comments}
\newacronym{rfr}{RFR}{Random Forest Regressor}
\newacronym{ric}{RIC}{RAN Intelligent Controller}
\newacronym{nrric}{Near-RT RIC}{Near-Real-Time \gls{ran} Intelligent Controller}
\newacronym{rlc}{RLC}{Radio Link Control}
\newacronym{rlf}{RLF}{Radio Link Failure}
\newacronym{rlnc}{RLNC}{Random Linear Network Coding}
\newacronym{rmr}{RMR}{RIC Message Router}
\newacronym{rmse}{RMSE}{Root Mean Squared Error}
\newacronym{rnis}{RNIS}{Radio Network Information Service}
\newacronym{rrc}{RRC}{Radio Resource Control}
\newacronym{rrm}{RRM}{Radio Resource Management}
\newacronym{rru}{RRU}{Remote Radio Unit}
\newacronym{rs}{RS}{Remote Server}
\newacronym{rsrp}{RSRP}{Reference Signal Received Power}
\newacronym{rsrq}{RSRQ}{Reference Signal Received Quality}
\newacronym{rss}{RSS}{Received Signal Strength}
\newacronym{rssi}{RSSI}{Received Signal Strength Indicator}
\newacronym{rtt}{RTT}{Round Trip Time}
\newacronym{ru}{RU}{Radio Unit}
\newacronym{rw}{RW}{Receive Window}
\newacronym{rx}{RX}{Receiver}
\newacronym{s1ap}{S1AP}{S1 Application Protocol}
\newacronym{sa}{SA}{standalone}
\newacronym{sack}{SACK}{Selective Acknowledgment}
\newacronym{sap}{SAP}{Service Access Point}
\newacronym{sc2}{SC2}{Spectrum Collaboration Challenge}
\newacronym{scef}{SCEF}{Service Capability Exposure Function}
\newacronym{sch}{SCH}{Secondary Cell Handover}
\newacronym{scoot}{SCOOT}{Split Cycle Offset Optimization Technique}
\newacronym{sctp}{SCTP}{Stream Control Transmission Protocol}
\newacronym{sdap}{SDAP}{Service Data Adaptation Protocol}
\newacronym{sdk}{SDK}{Software Development Kit}
\newacronym{sdm}{SDM}{Space Division Multiplexing}
\newacronym{sdma}{SDMA}{Spatial Division Multiple Access}
\newacronym{sdl}{SDL}{Shared Data Layer}
\newacronym{sdn}{SDN}{Software-defined Networking}
\newacronym{sdr}{SDR}{Software-defined Radio}
\newacronym{seba}{SEBA}{SDN-Enabled Broadband Access}
\newacronym{sgsn}{SGSN}{Serving GPRS Support Node}
\newacronym{sgw}{SGW}{Service Gateway}
\newacronym{si}{SI}{Study Item}
\newacronym{sib}{SIB}{Secondary Information Block}
\newacronym{sinr}{SINR}{Signal to Interference plus Noise Ratio}
\newacronym{sip}{SIP}{Session Initiation Protocol}
\newacronym{siso}{SISO}{Single Input, Single Output}
\newacronym{sla}{SLA}{Service Level Agreement}
\newacronym{sm}{SM}{Service Model}
\newacronym{e2sm}{E2SM}{E2 Service Model}
\newacronym{e2ap}{E2AP}{E2 Application Protocol}
\newacronym{smf}{SMF}{Session Management Function}
\newacronym{smo}{SMO}{Service Management and Orchestration}
\newacronym{sms}{SMS}{Short Message Service}
\newacronym{smsgmsc}{SMS-GMSC}{\gls{sms}-Gateway}
\newacronym{snr}{SNR}{Signal-to-Noise-Ratio}
\newacronym{son}{SON}{Self-Organizing Network}
\newacronym{sptcp}{SPTCP}{Single Path TCP}
\newacronym{srb}{SRB}{Service Radio Bearer}
\newacronym{srn}{SRN}{Standard Radio Node}
\newacronym{srs}{SRS}{Sounding Reference Signal}
\newacronym{ss}{SS}{Synchronization Signal}
\newacronym{sss}{SSS}{Secondary Synchronization Signal}
\newacronym{st}{ST}{Spanning Tree}
\newacronym{svc}{SVC}{Scalable Video Coding}
\newacronym{tb}{TB}{Transport Block}
\newacronym{tcp}{TCP}{Transmission Control Protocol}
\newacronym{tdd}{TDD}{Time Division Duplexing}
\newacronym{tdm}{TDM}{Time Division Multiplexing}
\newacronym{tdma}{TDMA}{Time Division Multiple Access}
\newacronym{tfl}{TfL}{Transport for London}
\newacronym{tfrc}{TFRC}{TCP-Friendly Rate Control}
\newacronym{tft}{TFT}{Traffic Flow Template}
\newacronym{tgen}{TGEN}{Traffic Generator}
\newacronym{tip}{TIP}{Telecom Infra Project}
\newacronym{tm}{TM}{Transparent Mode}
\newacronym{to}{TO}{Telco Operator}
\newacronym{tr}{TR}{Technical Report}
\newacronym{trp}{TRP}{Transmitter Receiver Pair}
\newacronym{ts}{TS}{Technical Specification}
\newacronym{tti}{TTI}{Transmission Time Interval}
\newacronym{ttt}{TTT}{Time-to-Trigger}
\newacronym{tx}{TX}{Transmitter}
\newacronym{uas}{UAS}{Unmanned Aerial System}
\newacronym{uav}{UAV}{Unmanned Aerial Vehicle}
\newacronym{udm}{UDM}{Unified Data Management}
\newacronym{udp}{UDP}{User Datagram Protocol}
\newacronym{udr}{UDR}{Unified Data Repository}
\newacronym{ue}{UE}{User Equipment}
\newacronym{uhd}{UHD}{\gls{usrp} Hardware Driver}
\newacronym{ul}{UL}{Uplink}
\newacronym{um}{UM}{Unacknowledged Mode}
\newacronym{uml}{UML}{Unified Modeling Language}
\newacronym{upa}{UPA}{Uniform Planar Array}
\newacronym{upf}{UPF}{User Plane Function}
\newacronym{urllc}{URLLC}{Ultra Reliable and Low Latency Communications}
\newacronym{usa}{U.S.}{United States}
\newacronym{usim}{USIM}{Universal Subscriber Identity Module}
\newacronym{usrp}{USRP}{Universal Software Radio Peripheral}
\newacronym{utc}{UTC}{Urban Traffic Control}
\newacronym{vim}{VIM}{Virtualization Infrastructure Manager}
\newacronym{vm}{VM}{Virtual Machine}
\newacronym{vnf}{VNF}{Virtual Network Function}
\newacronym{volte}{VoLTE}{Voice over \gls{lte}}
\newacronym{voltha}{VOLTHA}{Virtual OLT HArdware Abstraction}
\newacronym{vr}{VR}{Virtual Reality}
\newacronym{vran}{vRAN}{Virtualized RAN}
\newacronym{vss}{VSS}{Video Streaming Server}
\newacronym{wbf}{WBF}{Wired Bias Function}
\newacronym{wf}{WF}{Waterfilling}
\newacronym{wg}{WG}{Working Group}
\newacronym{wlan}{WLAN}{Wireless Local Area Network}
\newacronym{osm}{OSM}{Open Source \gls{nfv} Management and Orchestration}
\newacronym{pnf}{PNF}{Physical Network Function}
\newacronym{drl}{DRL}{Deep Reinforcement Learning}
\newacronym{mtc}{MTC}{Machine-type Communications}
\newacronym{osc}{OSC}{O-RAN Software Community}
\newacronym{mns}{MnS}{Management Services}
\newacronym{ves}{VES}{\gls{vnf} Event Stream}
\newacronym{ei}{EI}{Enrichment Information}
\newacronym{fh}{FH}{Fronthaul}
\newacronym{fft}{FFT}{Fast Fourier Transform}
\newacronym{laa}{LAA}{Licensed-Assisted Access}
\newacronym{plfs}{PLFS}{Physical Layer Frequency Signals}
\newacronym{ptp}{PTP}{Precision Time Protocol}
\newacronym{ntp}{NTP}{Network Time Protocol}
\newacronym{cbrs}{CBRS}{Citizen Broadband Radio Service}
\newacronym{rnti}{RNTI}{Radio Network Temporary Identifier}
\newacronym{tbs}{TBS}{Transport Block Size}
\newacronym{nfd}{NFD}{Node Feature Discovery}
\newacronym{mcp}{MCP}{Model Context Protocol}
\newacronym{vpn}{VPN}{Virtual Private Network}
\newacronym{onr}{ONR}{Office of Naval Research}
\newacronym{afosr}{AFOSR}{Air Force Office of Scientific Research}
\newacronym{afrl}{AFRL}{Air Force Research Laboratory}
\newacronym{arl}{ARL}{Army Research Laboratory}

\newacronym{arc}{ARC}{Aerial Research Cloud}

\newacronym{ct}{CT}{Continuous Testing}
\newacronym{mno}{MNO}{Mobile Network Operator}
\newacronym{oci}{OCI}{Open Container Initiative}
\newacronym{macsec}{MACsec}{Media Access Control Security}
\newacronym{pt}{PT}{Plain Text}
\newacronym{cuda}{CUDA}{Compute Unified Device Architecture}
\newacronym{dsp}{DSP}{Digital Signal Processing}

\newacronym{cus}{CUS}{Control, User, Synchronization}
\newacronym{dpd}{DPD}{Digital Pre-Distorsion}
\newacronym{cfr}{CFR}{Crest Factor Reduction}
\newacronym{pci}{PCIe}{Peripheral Component Interconnect Express}
\newacronym{dpu}{DPU}{Data Processing Unit}
\newacronym{rfsoc}{RFSoC}{Radio Frequency System-on-Chip}
\newacronym{if}{IF}{Intermediate Frequency}
\newacronym{nyu}{NYU}{New York University}
\newacronym{gh}{GH}{Grace Hopper}
\newacronym{trl}{TRL}{Technology Readiness Level}
\newacronym{srfa}{SRFA}{Special Research Focus Area}
\newacronym{qsfp}{QSFP}{quad small form factor pluggable}
\newacronym{pse}{PSE}{Performance Specialized Engine}
\newacronym{cae}{CAE}{Cognitive Analysis Engine}
\newacronym{simd}{SIMD}{Single Instruction/Multiple Data}
\newacronym{rt}{RT}{Real-Time}
\newacronym{nrt}{NRT}{Non-Real-Time}
\newacronym{asm}{ASM}{Advanced Sleep Mode}
\newacronym{aoa}{AoA}{Angle of Arrival}
\newacronym{eaxcid}{eAxC\_ID}{extended Antenna-Carrier Identifier}
\newacronym{bwp}{BWP}{Bandwidth Part}
\newacronym{dfe}{DFE}{Digital Front-End}
\newacronym{spi}{SPI}{Serial Peripheral Interface}
\newacronym{gpio}{GPIO}{General Purpose Input/Output}
\newacronym{nco}{NCO}{Numerically Controlled Oscillator}
\newacronym{lo}{LO}{Local Oscillator}
\newacronym{lna}{LNA}{Low-Noise Amplifier}
\newacronym{pll}{PLL}{Phased-Locked Loop}
\newacronym{som}{SOM}{System-on-Module}
\newacronym{papr}{PAPR}{Peak-to-Average Power Ratio}
\newacronym{pcb}{PCB}{Printed Circuit Board}
\newacronym{gcpw}{GCPW}{Grounded Co-Planar Waveguide}
\newacronym{cnn}{CNN}{Convolutional Neural Network}
\newacronym{gmp}{GMP}{Generalized Memory Polynomial}
\newacronym{ngrg}{nGRG}{next Generation Research Group}
\newacronym{mrl}{MRL}{Manufacturing Readiness Level}
\newacronym{fr}{FR}{Frequency Range}

\newacronym{sbom}{SBOM}{Software Bill of Materials}
\newacronym{hbom}{HBOM}{Hardware Bill of Materials}
\newacronym{vex}{VEX}{Vulnerability Exploitability eXchange}
\newacronym{dos}{DoS}{Denial of Service}
\newacronym{sme}{SME}{Small-Medium Enterprise}

\newacronym{ulpi}{ULPI}{Uplink Performance Improvement}
\newacronym{oem}{OEM}{Original Equipment Manufacturer}
\newacronym{nsin}{NSIN}{National Security Innovation Network}
\newacronym{dod}{DoD}{Department of Defense}
\newacronym{arpu}{ARPU}{Average Revenue per User}
\newacronym{opex}{OPEX}{operational expenses}
\newacronym{txb}{TXB}{Transmit Beam}
\newacronym{cve}{CVE}{Common Vulnerabilities and Exposure}
\newacronym{json}{JSON}{JavaScript Object Notation}
\newacronym{it}{IT}{Information Technology}
\newacronym{ci}{CI}{Continuous Integration}
\newacronym{nlp}{NLP}{Natural Language Processing}
\newacronym{nl}{NL}{Natural Language}
\newacronym{ulsch}{ULSCH}{Uplink Shared Agent}
\newacronym{phr}{PHR}{Power Headroom Report}
\newacronym{fwa}{FWA}{Fixed Wireless Access}
\newacronym{iq}{IQ}{In-Phase/Quadrature}
\newacronym{ram}{RAM}{Random Access Memory}
\newacronym{xr}{XR}{eXtended Reality}
\newacronym{aws}{AWS}{Amazon Web Services}
\newacronym{ra}{RA}{Resource Allocation}
\newacronym{pc}{PC}{Power Control}
\newacronym{a2a}{A2A}{Agent-To-Agent} 
\newacronym{icl}{ICL}{In-Context Learning} 
\newacronym{neu}{NU}{Northeastern University} 
\newacronym{sb}{SB}{SoftBank Corp., Japan}

%% file: figures/fwa_performance.tex
\begin{tikzpicture}
\pgfplotsset{every tick label/.append style={font=\scriptsize}}

\definecolor{darkgrey176}{RGB}{176,176,176}
\definecolor{green}{RGB}{0,128,0}
\definecolor{green01270}{RGB}{0,127,0}
\definecolor{lightgrey204}{RGB}{204,204,204}
\definecolor{purple}{RGB}{128,0,128}

\begin{axis}[
legend cell align={left},
legend style={
  fill opacity=0.8,
  draw opacity=1,
  text opacity=1,
  at={(2.25,1.05)},
  anchor=south,
  draw=lightgrey204,
  legend columns=5,
  inner xsep=0.9em,
  font=\footnotesize
},
width=\fwidth,
height=\fheight,
tick align=outside,
tick pos=left,
x grid style={darkgrey176},
xlabel={Time [s]},
xmajorgrids,
xmin=200, xmax=901.933,
xtick style={color=black},
y grid style={darkgrey176},
ylabel={Throughput [$\mathrm{Mbit/s}$]},
ymajorgrids,
ymin=0, ymax=104.85,
ytick style={color=black},
xlabel style={font=\footnotesize\color{white!15!black}},
ylabel style={font=\footnotesize\color{white!15!black}},
xlabel shift=-2pt,
ylabel shift=-5pt
]

\fill[red!10, opacity=0.5] (axis cs:308.288,0) rectangle (axis cs:605.356,104.85);
\node[fill=white, draw=black, anchor=south, font=\tiny, inner sep=1pt] at (axis cs:470,35) {Emergency};

\addplot [thick, blue]
table {%
-0.795 16.05
0.205 16.1571428571429
2.205 16.24375
3.205 16.2722222222222
4.205 16.055
5.205 16.1481818181818
6.205 16.0209090909091
7.205 16.0572727272727
8.205 15.5663636363636
9.205 15.6572727272727
11.205 15.5345454545455
12.205 15.7163636363636
13.205 15.7572727272727
14.205 15.7027272727273
15.205 15.6390909090909
16.205 15.8618181818182
17.205 15.8863636363636
18.205 16.0454545454545
19.205 16.0181818181818
20.205 16.4772727272727
21.205 16.4530303030303
22.205 17.3393939393939
23.205 17.530303030303
24.205 17.7984848484848
25.205 17.8727272727273
26.205 18.7181818181818
27.205 18.8181818181818
28.205 18.9136363636364
29.205 19.1454545454545
30.205 19.7272727272727
31.205 19.75
32.205 19.9333333333333
33.205 19.3106060606061
34.205 19.2787878787879
35.205 19.0666666666667
36.205 18.969696969697
37.205 18.6121212121212
38.205 18.5439393939394
39.205 18.6939393939394
40.205 18.5939393939394
41.205 18.3757575757576
42.205 18.2212121212121
43.205 18.7530303030303
44.205 19.0412121212121
45.205 18.9684848484848
47.205 19.2578787878788
48.205 19.6821212121212
49.205 19.139696969697
51.205 19.6851515151515
52.205 19.5624242424242
53.205 19.4715151515152
54.205 19.5836363636364
55.205 20.4290909090909
56.205 20.0381818181818
57.205 20.0484848484848
58.205 20.2075757575758
59.205 20.1424242424242
60.205 19.4363636363636
61.205 19.8030303030303
62.205 19.9484848484848
63.205 19.8787878787879
64.205 20.1878787878788
65.205 20.0848484848485
66.205 19.430303030303
67.205 19.3060606060606
68.205 19.2484848484848
69.205 19.5893939393939
70.205 19.4348484848485
71.205 20.3984848484848
72.205 20.5257575757576
73.205 20.2348484848485
74.205 20.4863636363636
75.205 20.6560606060606
76.205 21.0378787878788
77.205 21.4378787878788
78.205 21.55
79.205 21.5681818181818
80.205 21.1909090909091
81.205 21.8151515151515
82.205 21.3363636363636
83.205 20.9515151515152
85.205 20.8833333333333
86.205 20.75
87.205 20.9075757575758
88.205 20.65
89.205 20.7136363636364
90.205 20.5212121212121
91.205 20.1181818181818
92.205 20.0090909090909
93.205 19.5151515151515
94.205 19.6484848484848
95.205 19.930303030303
96.205 20.080303030303
97.205 20.180303030303
98.205 19.7348484848485
99.205 19.6378787878788
100.205 19.419696969697
101.205 19.6151515151515
102.205 19.9818181818182
103.205 19.7727272727273
104.205 19.8181818181818
105.205 19.5636363636364
106.205 19.7606060606061
107.205 19.5242424242424
108.205 19.530303030303
109.205 19.880303030303
110.205 19.7530303030303
111.205 19.7681818181818
112.205 19.9772727272727
113.205 19.8909090909091
114.205 20.4090909090909
115.205 20.2757575757576
116.205 20.4030303030303
117.205 19.9424242424242
118.205 19.9924242424242
119.205 20.080303030303
120.205 19.2484848484848
121.205 19.3
122.205 19.3575757575758
123.205 19.2666666666667
124.205 19.3257575757576
125.205 19.0075757575758
126.205 19.1772727272727
127.205 19.2227272727273
128.205 19.7318181818182
129.205 19.5772727272727
130.205 19.8348484848485
131.205 20.4772727272727
132.205 20.3984848484848
133.205 20.2257575757576
134.205 20.2712121212121
135.205 20.2106060606061
136.205 20.4560606060606
137.205 20.3287878787879
138.205 20.8015151515152
139.205 20.5833333333333
140.205 20.8606060606061
141.205 20.3333333333333
142.205 20.2787878787879
143.205 20.0424242424242
144.205 20.1787878787879
145.205 19.6606060606061
146.205 19.7393939393939
147.205 19.6757575757576
148.205 19.5878787878788
149.205 18.869696969697
150.205 18.8424242424242
151.205 18.7333333333333
152.205 18.8924242424242
153.205 19.0090909090909
154.205 19.4045454545455
155.205 19.3469696969697
157.205 19.3924242424242
158.205 19.3742424242424
159.205 19.0469696969697
160.205 19.2893939393939
161.205 19.6257575757576
162.205 19.6166666666667
163.205 19.8257575757576
164.205 19.8212121212121
165.205 19.7575757575758
166.205 19.3621212121212
167.205 19.3378787878788
168.205 19.6378787878788
169.205 19.6348484848485
170.205 19.8893939393939
171.205 19.7833333333333
172.205 19.7560606060606
173.205 19.8060606060606
174.205 19.4060606060606
175.205 19.5469696969697
176.205 19.3878787878788
177.205 19.7545454545455
178.205 19.6727272727273
179.205 19.6454545454545
180.205 19.4060606060606
181.205 19.5606060606061
182.205 19.6015151515152
183.205 19.5227272727273
184.205 19.8
185.205 19.9272727272727
186.205 19.45
187.205 19.3681818181818
188.205 19.3469696969697
189.205 19.3469696969697
190.205 19.0015151515152
191.205 19.3469696969697
193.205 19.019696969697
194.205 19.1212121212121
195.205 18.8272727272727
196.205 18.3272727272727
197.205 18.2272727272727
198.205 18.2363636363636
199.205 18.4606060606061
200.205 18.2242424242424
201.205 18.5787878787879
202.205 18.7181818181818
203.205 18.4606060606061
204.205 18.8212121212121
205.205 18.3515151515152
206.205 18.9121212121212
207.205 19.030303030303
208.205 18.8212121212121
209.205 19.2787878787879
210.205 19.1090909090909
211.205 19.6090909090909
212.205 19.2969696969697
213.205 19.7939393939394
214.205 20.0090909090909
215.205 19.9575757575758
216.205 20.369696969697
217.205 20.3454545454545
218.205 20.2909090909091
219.205 20.7590909090909
220.205 20.9833333333333
221.205 20.8378787878788
222.205 20.6333333333333
223.205 20.4090909090909
224.205 19.9636363636364
225.205 19.8878787878788
226.205 19.5333333333333
227.205 19.5757575757576
228.205 19.3121212121212
229.205 19.4030303030303
230.205 19.2893939393939
231.205 19.0257575757576
232.205 19.130303030303
233.205 18.9530303030303
235.205 19.5030303030303
238.205 19.5689393939394
239.205 19.7659090909091
240.205 19.9780303030303
241.205 19.9871212121212
242.205 19.705303030303
244.205 20.2280303030303
245.205 19.9416666666667
246.205 20.2189393939394
247.205 20.2962121212121
248.205 20.3719696969697
249.205 20.0537878787879
250.205 20.0424242424242
251.205 19.8924242424242
253.205 19.9712121212121
254.205 19.8166666666667
255.205 20.1098484848485
257.205 20.0643939393939
258.205 20.1916666666667
259.205 20.0780303030303
260.205 19.8871212121212
262.205 19.9143939393939
263.205 20.155303030303
264.205 20.0734848484848
265.205 20.0325757575758
266.205 20.1871212121212
267.205 20.1962121212121
268.205 20.4484848484848
269.205 20.0075757575758
270.205 20.1575757575758
271.205 19.8969696969697
272.205 20.0060606060606
273.205 19.9848484848485
274.205 19.8636363636364
275.205 19.9272727272727
277.205 19.9545454545455
278.205 19.6393939393939
281.205 19.7166666666667
282.205 19.630303030303
283.205 19.530303030303
284.205 19.5848484848485
285.205 19.64
286.205 19.8331818181818
287.205 19.8581818181818
288.205 20.0127272727273
289.205 20.1581818181818
291.205 20.1781818181818
293.205 20.4706060606061
295.205 20.4533333333333
296.205 20.1715151515152
297.205 20.4624242424242
299.205 20.4942424242424
303.205 20.4512121212121
305.205 20.2671212121212
306.205 20.2057575757576
307.205 20.2451515151515
308.205 20.0451515151515
309.205 20.2387878787879
310.205 20.0281818181818
311.205 20.0469696969697
312.205 19.9742424242424
313.205 19.8924242424242
314.205 19.369696969697
315.205 19.1318181818182
317.205 19.0045454545455
318.205 19.1227272727273
319.205 19.15
320.205 19.2166666666667
321.205 18.969696969697
322.205 19.0530303030303
323.205 19.069696969697
324.205 19.4742424242424
325.205 19.419696969697
327.205 19.869696969697
328.205 20.2515151515152
329.205 20.3924242424242
330.205 20.4515151515152
331.205 20.0787878787879
332.205 20.2484848484848
333.205 20.2681818181818
334.205 20.6181818181818
335.205 20.5166666666667
336.205 20.5121212121212
337.205 20.2848484848485
338.205 20.4530303030303
339.205 20.0848484848485
340.205 20.6075757575758
341.205 20.2424242424242
342.205 20.4242424242424
343.205 20.5015151515152
344.205 20.3333333333333
345.205 20.1151515151515
346.205 19.8787878787879
347.205 20.2469696969697
348.205 20.219696969697
349.205 20.3742424242424
350.205 20.5742424242424
351.205 20.4651515151515
352.205 20.9530303030303
353.205 20.9227272727273
354.205 20.8363636363636
355.205 20.9151515151515
356.205 20.069696969697
357.205 19.4933333333333
358.205 18.2074242424242
359.205 17.8528787878788
360.205 16.7731818181818
361.205 15.9513636363636
362.205 14.9877272727273
363.205 13.9513636363636
364.205 13.2089393939394
365.205 12.3725757575758
366.205 11.6907575757576
367.205 11.6543939393939
368.205 11.7277272727273
369.205 11.4981818181818
370.205 10.7281818181818
371.205 10.034696969697
372.205 9.62924242424243
373.205 8.99606060606061
374.205 8.37484848484849
375.205 7.65121212121212
376.205 7.07757575757576
377.205 6.39424242424242
378.205 5.67878787878788
379.205 5.05969696969697
380.205 4.80287878787879
381.205 4.79924242424242
382.205 4.86242424242424
383.205 4.72106060606061
384.205 4.85393939393939
385.205 5.90242424242424
386.205 6.97454545454545
387.205 8.07545454545455
388.205 9.09818181818182
389.205 10.4954545454545
390.205 11.3569696969697
391.205 12.6201515151515
392.205 13.5816666666667
393.205 14.8371212121212
394.205 15.8118181818182
395.205 16.7030303030303
396.205 16.7333333333333
397.205 17.030303030303
398.205 16.7212121212121
399.205 15.975303030303
400.205 14.7007575757576
401.205 13.8165151515152
402.205 12.4892424242424
403.205 11.6681818181818
404.205 10.3536363636364
405.205 9.3830303030303
406.205 8.33348484848485
407.205 7.28636363636364
408.205 5.95090909090909
409.205 5.08745454545455
410.205 4.74109090909091
411.205 4.62518181818182
412.205 4.84018181818182
413.205 5.77654545454546
415.205 6.16427272727273
416.205 7.24548484848485
417.205 8.28275757575757
418.205 8.99593939393939
419.205 10.0218484848485
420.205 10.9936666666667
421.205 11.7116666666667
422.205 12.8524242424242
423.205 13.8519696969697
424.205 14.6742424242424
425.205 14.8106060606061
426.205 14.9833333333333
427.205 14.495303030303
428.205 14.0534848484848
429.205 14.0807575757576
430.205 13.6898484848485
431.205 13.240303030303
432.205 13.1093939393939
433.205 12.67
434.205 12.4245454545455
435.205 12.0018181818182
436.205 11.5518181818182
437.205 11.4427272727273
438.205 11.5007575757576
439.205 11.3516666666667
440.205 11.3198484848485
441.205 11.2925757575758
442.205 11.3875757575758
443.205 11.2457575757576
444.205 11.1012121212121
445.205 10.7844848484848
446.205 10.571303030303
447.205 10.2949393939394
448.205 9.8699393939394
449.205 9.6899393939394
451.205 9.83539393939394
452.205 9.51857575757576
453.205 9.30372727272727
454.205 9.071
455.205 9.16190909090909
456.205 9.11872727272727
457.205 9.21060606060606
458.205 9.25833333333333
459.205 9.56045454545455
460.205 9.99909090909091
461.205 10.0787878787879
462.205 10.1818181818182
463.205 10.5395454545455
464.205 11.1225757575758
465.205 11.4916666666667
466.205 11.5098484848485
467.205 12.0430303030303
468.205 12.1587878787879
469.205 12.4939393939394
470.205 12.7272727272727
471.205 12.7590909090909
472.205 13.1257575757576
473.205 13.0863636363636
474.205 13.0712121212121
475.205 12.9757575757576
476.205 12.7621212121212
477.205 12.689696969697
478.205 12.2715151515151
479.205 12.3124242424242
480.205 12.0942424242424
481.205 11.7760606060606
482.205 11.7806060606061
483.205 11.4360606060606
484.205 11.3906060606061
485.205 11.3466666666667
486.205 11.0193939393939
487.205 11.2330303030303
488.205 11.4372727272727
489.205 11.4033333333333
490.205 11.209696969697
491.205 11.0759090909091
492.205 10.954696969697
493.205 10.6474242424242
494.205 10.4074242424242
495.205 10.0715151515152
496.205 9.88515151515152
497.205 9.8030303030303
498.205 9.50712121212121
499.205 9.23075757575758
500.205 9.21560606060606
501.205 9.14439393939394
502.205 9.06393939393939
503.205 8.78424242424242
504.205 8.71878787878788
505.205 8.76939393939394
506.205 8.69712121212121
507.205 8.40590909090909
508.205 8.25075757575758
509.205 8.10530303030303
510.205 7.96484848484848
511.205 7.79363636363636
512.205 7.72757575757576
513.205 7.63181818181818
514.205 7.57636363636364
515.205 7.57727272727273
516.205 7.32181818181818
517.205 7.20575757575757
518.205 7.29606060606061
519.205 7.07878787878788
521.205 6.85651515151515
522.205 6.78606060606061
523.205 6.65181818181818
524.205 6.59181818181818
525.205 6.41515151515152
526.205 6.31787878787879
527.205 6.17560606060606
528.205 6.30015151515151
529.205 6.37348484848485
530.205 6.15166666666667
531.205 6.25348484848485
532.205 6.26893939393939
533.205 6.14621212121212
534.205 6.00257575757576
535.205 5.83212121212121
536.205 5.95
537.205 5.94954545454545
538.205 5.91
539.205 5.73272727272727
540.205 5.66818181818182
541.205 5.6
542.205 5.51030303030303
543.205 5.57121212121212
544.205 5.67393939393939
545.205 5.77984848484848
546.205 5.81393939393939
547.205 5.77121212121212
548.205 5.89893939393939
549.205 5.81893939393939
550.205 5.87575757575758
551.205 5.90121212121212
552.205 6.04984848484848
553.205 5.99863636363636
554.205 5.91409090909091
555.205 5.81909090909091
556.205 5.73
557.205 5.66181818181818
558.205 5.56969696969697
559.205 5.38727272727273
560.205 5.382
561.205 5.36290909090909
562.205 5.33881818181818
563.205 5.18972727272727
564.205 5.047
567.205 4.96427272727273
568.205 4.74290909090909
569.205 4.70109090909091
570.205 4.82927272727273
571.205 4.84412121212121
572.205 4.80715151515151
573.205 4.78242424242424
574.205 4.60015151515152
575.205 4.56424242424242
576.205 4.72560606060606
577.205 4.80272727272727
578.205 4.87181818181818
579.205 5
580.205 5.03742424242424
581.205 4.91742424242424
584.205 4.93106060606061
585.205 5.035
586.205 5.04772727272727
587.205 5.17621212121212
588.205 5.12621212121212
589.205 4.96939393939394
590.205 4.98318181818182
591.205 5.07272727272727
593.205 5.06757575757576
594.205 4.97666666666667
595.205 4.84939393939394
596.205 4.88393939393939
597.205 4.8430303030303
598.205 4.85212121212121
599.205 4.88333333333333
600.205 4.87606060606061
601.205 4.88515151515152
602.205 4.85151515151515
603.205 4.62530303030303
605.205 4.665
606.205 4.80621212121212
607.205 5.01075757575758
608.205 4.86530303030303
609.205 4.86984848484849
610.205 4.90348484848485
611.205 4.90960606060606
612.205 4.97506060606061
613.205 5.04324242424242
614.205 5.0684696969697
616.205 5.14713636363636
617.205 5.11077272727273
618.205 4.99213636363636
619.205 5.92031818181818
621.205 6.85940909090909
624.205 8.09213636363636
626.205 9.09057792207792
629.205 10.2629415584416
630.205 11.2714264069264
631.205 12.46279004329
634.205 13.23029004329
635.205 14.4222294372294
636.205 15.2967748917749
637.205 16.732683982684
638.205 16.5917748917749
639.205 16.7235930735931
640.205 16.4417748917749
641.205 16.9424242424242
642.205 17.4
643.205 17.7151515151515
644.205 17.9969696969697
645.205 19.0333333333333
646.205 19.6212121212121
647.205 20.5893939393939
648.205 20.8530303030303
649.205 21.9257575757576
650.205 22.7075757575758
651.205 23.1621212121212
652.205 23.4121212121212
653.205 23.5636363636364
654.205 24.2545454545455
655.205 24.6454545454545
656.205 24.4833333333333
657.205 24.4878787878788
658.205 24.4015151515151
659.205 24.219696969697
660.205 24.2742424242424
661.205 24.4969696969697
662.205 24.7151515151515
663.205 24.6969696969697
664.205 24.6060606060606
665.205 24.4151515151515
666.205 23.8469696969697
667.205 23.9075757575758
668.205 23.7166666666667
669.205 23.6348484848485
670.205 23.8757575757576
671.205 23.4621212121212
672.205 23.3181818181818
673.205 23.5227272727273
674.205 23.45
675.205 23.7545454545455
676.205 23.55
677.205 23.5181818181818
678.205 23.1242424242424
680.205 23.1106060606061
681.205 22.7833333333333
682.205 22.6909090909091
683.205 22.3636363636364
684.205 22.080303030303
685.205 21.6393939393939
686.205 21.4636363636364
687.205 21.2863636363636
688.205 21.1787878787879
689.205 21.6242424242424
690.205 21.9742424242424
691.205 22.3015151515151
692.205 22.4681818181818
693.205 22.4651515151515
694.205 22.6545454545455
695.205 22.2227272727273
696.205 22.2136363636364
697.205 21.9833333333333
698.205 21.3378787878788
699.205 21.0439393939394
700.205 20.680303030303
701.205 20.2848484848485
702.205 19.8575757575758
703.205 19.7318181818182
704.205 19.4924242424242
705.205 19.1530303030303
706.205 19.2166666666667
707.205 19.5166666666667
708.205 19.4909090909091
709.205 19.7772727272727
710.205 19.9045454545455
711.205 20.0863636363636
712.205 20.0909090909091
713.205 20.2
714.205 20.0090909090909
715.205 20.1621212121212
717.205 20.4378787878788
718.205 20.4469696969697
719.205 19.9651515151515
720.205 20.0181818181818
721.205 20.0863636363636
722.205 20.0651515151515
723.205 19.9060606060606
724.205 19.9515151515152
725.205 19.7515151515152
726.205 20.0166666666667
727.205 19.3575757575758
728.205 19.4151515151515
729.205 19.1060606060606
730.205 19.430303030303
731.205 19.1545454545455
732.205 19.1090909090909
733.205 18.9727272727273
734.205 18.3590909090909
735.205 18.7636363636364
736.205 18.6
737.205 18.5939393939394
738.205 18.9515151515151
739.205 18.9545454545455
740.205 19.3969696969697
741.205 19.0727272727273
742.205 19.5
743.205 19.2666666666667
744.205 19.3621212121212
745.205 19.4530303030303
746.205 19.5348484848485
747.205 19.5045454545455
748.205 19.5712121212121
749.205 19.2045454545455
750.205 19.2469696969697
751.205 18.9681818181818
752.205 18.95
753.205 18.9530303030303
754.205 18.55
755.205 18.7121212121212
756.205 18.6939393939394
757.205 18.4212121212121
758.205 18.3424242424242
759.205 18.1424242424242
760.205 18.4969696969697
761.205 18.3863636363636
762.205 18.6257575757576
763.205 18.4893939393939
764.205 18.4560606060606
765.205 18.3378787878788
766.205 18.3166666666667
767.205 18.5772727272727
768.205 18.5045454545455
769.205 18.7590909090909
770.205 18.4287878787879
771.205 18.8787878787879
772.205 18.6439393939394
773.205 18.5045454545455
774.205 18.8954545454545
775.205 18.780303030303
776.205 18.880303030303
777.205 18.7439393939394
778.205 18.719696969697
779.205 18.4893939393939
780.205 18.5530303030303
781.205 18.8939393939394
782.205 18.5954545454545
783.205 18.8439393939394
784.205 19.1075757575758
785.205 19.1766666666667
786.205 19.1130303030303
787.205 19.5675757575758
788.205 19.6766666666667
789.205 19.9584848484848
791.205 20.3706060606061
792.205 20.2160606060606
793.205 20.7206060606061
794.205 20.7690909090909
795.205 20.7418181818182
796.205 20.5842424242424
797.205 20.0742424242424
798.205 20.3469696969697
799.205 20.3166666666667
800.205 19.5257575757576
801.205 19.4257575757576
802.205 19.3257575757576
803.205 19.2893939393939
804.205 18.3166666666667
805.205 18.2015151515152
806.205 18.1287878787879
807.205 18.0833333333333
808.205 18.6151515151515
809.205 18.169696969697
810.205 18.2681818181818
811.205 18.7454545454545
812.205 18.8787878787879
813.205 18.7333333333333
814.205 18.8393939393939
815.205 19.6212121212121
816.205 19.7636363636364
817.205 19.8863636363636
818.205 19.9984848484848
819.205 20.1484848484848
820.205 20.5348484848485
821.205 20.4757575757576
822.205 20.7530303030303
823.205 20.9787878787879
824.205 20.9060606060606
826.205 21.1030303030303
827.205 20.4666666666667
828.205 20.280303030303
829.205 20.2393939393939
830.205 20.2712121212121
831.205 19.5575757575758
832.205 19.3318181818182
833.205 19.0136363636364
834.205 19.0469696969697
835.205 18.5151515151515
836.205 18.7060606060606
837.205 18.6030303030303
838.205 18.8484848484848
839.205 19.380303030303
840.205 19.3075757575758
841.205 19.7575757575758
842.205 20.280303030303
843.205 20.4954545454545
844.205 20.7409090909091
845.205 20.6833333333333
846.205 21.1742424242424
847.205 21.2742424242424
848.205 21.219696969697
849.205 21.3651515151515
850.205 21.0893939393939
851.205 21.2893939393939
852.205 21.1166666666667
853.205 20.9393939393939
854.205 21.1909090909091
855.205 21.2121212121212
856.205 20.8545454545455
857.205 20.7181818181818
858.205 20.6772727272727
859.205 20.6075757575758
860.205 20.2348484848485
861.205 20.519696969697
862.205 20.25
863.205 19.8090909090909
864.205 19.7727272727273
865.205 19.6227272727273
866.205 19.5681818181818
867.205 20.0227272727273
868.205 19.7409090909091
869.205 19.6227272727273
870.205 19.2924242424242
871.205 19.6772727272727
872.205 19.1954545454545
873.205 19.3712121212121
874.205 19.6075757575758
875.205 19.3984848484848
876.205 19.4984848484848
877.205 19.480303030303
878.205 18.9712121212121
879.205 19.2772727272727
880.205 18.6818181818182
881.205 19.2515151515152
882.205 19.0212121212121
883.205 19.1393939393939
884.205 18.8242424242424
885.205 18.7954545454545
886.205 19.2590909090909
887.205 18.9772727272727
888.205 19.1151515151515
889.205 19.169696969697
890.205 19.2181818181818
891.205 19.4363636363636
892.205 19.4
893.205 19.2818181818182
894.205 19.0909090909091
895.205 19.3363636363636
896.205 19.0606060606061
897.205 18.8366666666667
898.205 18.6907407407407
899.205 18.5708333333333
900.205 18.8809523809524
901.205 18.3277777777778
};
\addlegendentry{Average Achieved Throughput}

\addplot [very thick, red, const plot mark left, dotted]
table {%
0 100
13.096 100
328.543 60
341.84 25
356.642 10
372.244 3
387.817 15
400.945 3
416.102 15
429.233 10
443.406 8
461.871 12
476.016 10
492.186 8
505.57 6
517.477 5
533.861 4
559.119 3
622.102 15
640.937 25
659.593 35
672.893 40
687.853 50
702.811 60
716.316 35
730.48 40
744.879 60
759.053 55
772.224 50
787.438 45
802.795 35
816.748 60
830.86 50
844.847 70
865.347 50
879.599 65
893.551 80
};
\addlegendentry{Throughput Throttling Limit}

\addplot [purple, opacity=0.7, densely dashdotted, forget plot, very thick]
table {%
308.288 1.77635683940025e-15
308.288 104.85
};
\addplot [purple, opacity=0.7, densely dashdotted, forget plot, very thick]
table {%
605.356 1.77635683940025e-15
605.356 104.85
};

\addlegendimage{purple, densely dashdotted, very thick, opacity=0.7}
\addlegendentry{Intent Change}

\addlegendimage{green, dashed, very thick, opacity=0.7}
\addlegendentry{Target SNR}

\addlegendimage{area legend, fill=green, fill opacity=0.3, draw=green, thick}
\addlegendentry{Power}

\end{axis}

\begin{axis}[
width=\fwidth,
height=\fheight,
axis y line*=right,
tick align=outside,
x grid style={darkgrey176},
xtick=\empty,  
xmin=200, xmax=901.933,
xtick pos=left,
xtick style={color=black},
y grid style={darkgrey176},
ylabel=\textcolor{green}{Target SNR [dB]},
ymin=0, ymax=19,
ytick pos=right,
ytick style={color=green},
yticklabel style={anchor=west},
ylabel style={font=\footnotesize\color{white!15!black}},
ylabel shift=-4pt
]
\addplot [very thick, green01270, const plot mark left, dashed]
table {%
0 18
4.515 18
22.714 18
43.618 18
63.878 18
80.862 18
100.344 18
118.337 18
137.375 18
155.581 18
176.273 18
196.899 18
214.707 18
235.567 18
258.666 18
282.159 18
311.749 18
333.864 18
356.027 18
372.666 15
390.696 15
408.731 12
426.163 15
439.918 12
456.946 16.5
475.402 15
494.895 18
511.521 18
530.393 18
551.318 18
572.596 18
597.562 18
618.436 15
640.732 18
661.426 18
679.64 18
696.448 18
711.812 18
730.685 18
746.725 18
763.173 18
779.416 18
795.415 18
810.78 18
828.206 18
844.429 18
862.698 18
880.008 18
};

\end{axis}
\end{tikzpicture}

%% file: figures/mtc_performance.tex
\begin{tikzpicture}
\pgfplotsset{every tick label/.append style={font=\scriptsize}}

\definecolor{darkgrey176}{RGB}{176,176,176}
\definecolor{green}{RGB}{0,128,0}
\definecolor{green01270}{RGB}{0,127,0}
\definecolor{lightgrey204}{RGB}{204,204,204}
\definecolor{purple}{RGB}{128,0,128}

\begin{axis}[
legend cell align={left},
legend style={
  fill opacity=0.8,
  draw opacity=1,
  text opacity=1,
  at={(2.25,1.05)},
  anchor=south,
  draw=lightgrey204,
  legend columns=5,
  inner xsep=0.9em,
  font=\footnotesize
},
width=\fwidth,
height=\fheight,
tick align=outside,
tick pos=left,
x grid style={darkgrey176},
xlabel={Time [s]},
xmajorgrids,
xmin=200, xmax=901.933,
xtick style={color=black},
y grid style={darkgrey176},
ylabel={Throughput [$\mathrm{Mbit/s}$]},
ymajorgrids,
ymin=0, ymax=104.894136363636,
ytick style={color=black},
xlabel style={font=\footnotesize\color{white!15!black}},
ylabel style={font=\footnotesize\color{white!15!black}},
xlabel shift=-2pt,
ylabel shift=-5pt
]

\fill[red!10, opacity=0.5] (axis cs:308.288,0) rectangle (axis cs:605.356,180);
\node[fill=white, draw=black, anchor=south, font=\tiny, inner sep=1pt] at (axis cs:470,55) {Emergency};

\addplot [thick, blue]
table {%
0.205 12.225
2.205 12.35
3.205 12.43125
4.205 12.3522222222222
5.205 12.287
7.205 12.2336363636364
8.205 12.3790909090909
9.205 12.0827272727273
11.205 12.0009090909091
12.205 12.0372727272727
13.205 12.3736363636364
14.205 12.5327272727273
15.205 12.5509090909091
16.205 12.36
17.205 12.7945454545455
18.205 12.9945454545455
19.205 13.14
20.205 13.0218181818182
21.205 13.1181818181818
22.205 13.2
23.205 13.2545454545455
24.205 13.0545454545455
25.205 13.4818181818182
26.205 13.9090909090909
27.205 14.5
28.205 15.2272727272727
29.205 15.6
30.205 16.1909090909091
31.205 16.1363636363636
32.205 17.3363636363636
33.205 17.9
34.205 17.9090909090909
35.205 18.7909090909091
36.205 18.8818181818182
37.205 18.6
38.205 18.3181818181818
39.205 18.0272727272727
40.205 17.9181818181818
41.205 17.6772727272727
42.205 17.7136363636364
43.205 17.2045454545455
44.205 17.0272727272727
45.205 17.2181818181818
47.205 16.7181818181818
48.205 16.2909090909091
49.205 16.5
51.205 16.4
52.205 16.1090909090909
53.205 16.1818181818182
54.205 16.2681818181818
55.205 16.2318181818182
56.205 16.1681818181818
57.205 16.0090909090909
58.205 16.0818181818182
59.205 15.7727272727273
60.205 16.3
61.205 16.5727272727273
62.205 17.0545454545455
63.205 17.2090909090909
64.205 16.8636363636364
65.205 16.8
66.205 17.2636363636364
67.205 17.4454545454545
68.205 17.5363636363636
69.205 17.7545454545455
70.205 17.7090909090909
71.205 17.5272727272727
72.205 16.9727272727273
73.205 16.8454545454545
74.205 16.2363636363636
75.205 16.7454545454545
76.205 16.3818181818182
77.205 16.4272727272727
78.205 16.1363636363636
79.205 16.05
80.205 16.0409090909091
81.205 16.3772727272727
82.205 16.15
83.205 16.1772727272727
85.205 16.5954545454545
86.205 16.9045454545455
87.205 16.6409090909091
88.205 16.8136363636364
89.205 16.75
90.205 16.4954545454545
91.205 16.8818181818182
92.205 16.8909090909091
93.205 17.1272727272727
94.205 17.6636363636364
95.205 17.8090909090909
96.205 17.3090909090909
97.205 17.1181818181818
98.205 17.4454545454545
99.205 17.3
100.205 17.4454545454545
101.205 17.5727272727273
102.205 17.5636363636364
103.205 17.0636363636364
104.205 17.0272727272727
105.205 16.1454545454545
106.205 16.6272727272727
107.205 16.2363636363636
108.205 16.9727272727273
109.205 16.0909090909091
110.205 16.9090909090909
111.205 16.2818181818182
112.205 16.6454545454545
113.205 16.1181818181818
114.205 16.6181818181818
115.205 16.2545454545455
116.205 16.6090909090909
117.205 16.4545454545455
118.205 16.8818181818182
119.205 16.2
120.205 16.2090909090909
121.205 15.9090909090909
122.205 16.0636363636364
123.205 15.9181818181818
124.205 15.9363636363636
125.205 15.8090909090909
126.205 15.7909090909091
127.205 15.9181818181818
128.205 15.4727272727273
129.205 15.6363636363636
130.205 15.3181818181818
131.205 16.1454545454545
132.205 15.8090909090909
133.205 16.3727272727273
134.205 16.6
135.205 17.1272727272727
136.205 17.2636363636364
137.205 17.8545454545455
138.205 18.4454545454545
139.205 18.2
140.205 17.9636363636364
141.205 18.6636363636364
142.205 18.5636363636364
143.205 18.5181818181818
144.205 17.9909090909091
145.205 17.8181818181818
146.205 17.3181818181818
147.205 17.0272727272727
148.205 16.6
149.205 16.3818181818182
150.205 16.8181818181818
151.205 16.8318181818182
152.205 16.7409090909091
153.205 16.2136363636364
154.205 16.4863636363636
155.205 16.8772727272727
157.205 17.3681818181818
158.205 17.3409090909091
159.205 17.5409090909091
160.205 17.2954545454545
161.205 16.9863636363636
162.205 16.7863636363636
163.205 17.0454545454545
164.205 16.7
165.205 17.2454545454545
166.205 17.0727272727273
167.205 16.7545454545455
168.205 16.3090909090909
169.205 16.4181818181818
170.205 16.4181818181818
171.205 16.4454545454545
172.205 16.4090909090909
173.205 16.3636363636364
174.205 16.3545454545455
175.205 16.3272727272727
176.205 16.5636363636364
177.205 16.2727272727273
178.205 16.5545454545455
179.205 16.2818181818182
180.205 16.8090909090909
181.205 16.9363636363636
182.205 17.1363636363636
183.205 16.9363636363636
184.205 17.2909090909091
185.205 16.9818181818182
186.205 17.0818181818182
187.205 16.95
188.205 17.7227272727273
189.205 17.7136363636364
190.205 18.0318181818182
191.205 17.4772727272727
193.205 17.35
194.205 17.8045454545455
195.205 18.0954545454545
196.205 18.6681818181818
197.205 18.8227272727273
198.205 19.1136363636364
199.205 19.1272727272727
200.205 18.5545454545455
201.205 18.8090909090909
202.205 18.5545454545455
203.205 18.7545454545455
204.205 18.9727272727273
205.205 18.3636363636364
206.205 18.4363636363636
207.205 17.5363636363636
208.205 17.8090909090909
209.205 17.5
210.205 17.7090909090909
211.205 17.9818181818182
212.205 17.9545454545455
213.205 18.6
214.205 18.4272727272727
215.205 17.9818181818182
216.205 18.0454545454545
217.205 17.8818181818182
218.205 18.3272727272727
219.205 18.2272727272727
220.205 18.8545454545455
221.205 18.2818181818182
222.205 18.6909090909091
223.205 18.5727272727273
224.205 18.4590909090909
225.205 18.6681818181818
226.205 19.2136363636364
227.205 19.3045454545455
228.205 19.5454545454545
230.205 19.2727272727273
231.205 19.0393939393939
232.205 18.5666666666667
233.205 18.3212121212121
235.205 17.9575757575758
236.205 17.5348484848485
239.205 16.9757575757576
240.205 16.7393939393939
241.205 16.230303030303
242.205 16.430303030303
244.205 15.9621212121212
245.205 15.9439393939394
246.205 15.9318181818182
247.205 15.8590909090909
248.205 16.2772727272727
249.205 16.5045454545455
250.205 17.0636363636364
251.205 17.7363636363636
253.205 18.1090909090909
254.205 18.2454545454545
255.205 18.3727272727273
256.205 18.9636363636364
257.205 19.2545454545455
258.205 19.4
259.205 19.5636363636364
260.205 19.6909090909091
262.205 20.0636363636364
263.205 20
264.205 19.7545454545455
265.205 19.6545454545455
266.205 19.3454545454545
267.205 19.5
268.205 18.7909090909091
269.205 18.8
270.205 18.8818181818182
271.205 18.8909090909091
272.205 18.9272727272727
273.205 18.0757575757576
274.205 17.5939393939394
275.205 17.4212121212121
277.205 17.1848484848485
278.205 17.6757575757576
281.205 17.630303030303
282.205 18.330303030303
283.205 17.8575757575758
284.205 17.7363636363636
285.205 17.5909090909091
286.205 17.4242424242424
287.205 17.5666666666667
288.205 17.4939393939394
291.205 17.4412121212121
293.205 17.5381818181818
295.205 17.4018181818182
296.205 17.1518181818182
297.205 17.1154545454545
303.205 17.7972727272727
305.205 17.9548484848485
306.205 17.9639393939394
308.205 17.8578787878788
309.205 18.2124242424242
310.205 18.1851515151515
311.205 18.0878787878788
312.205 18.8
313.205 18.8818181818182
314.205 18.85
315.205 18.35
317.205 18.5863636363636
318.205 18.4227272727273
319.205 18.3863636363636
320.205 18.75
321.205 18.3045454545455
322.205 18.95
323.205 19.5545454545455
324.205 18.9272727272727
325.205 18.9454545454545
326.205 19.1090909090909
327.205 19.3545454545455
328.205 19.4727272727273
329.205 19.5090909090909
330.205 20.2727272727273
331.205 20.2636363636364
332.205 20.5818181818182
333.205 20.2090909090909
334.205 19.3636363636364
335.205 19.4090909090909
336.205 19.1818181818182
337.205 19.4454545454545
338.205 19.7
339.205 18.9
340.205 19.7818181818182
341.205 19.2636363636364
342.205 18.7909090909091
343.205 18.3545454545455
344.205 18.5272727272727
345.205 19.4454545454545
346.205 19.5090909090909
347.205 19.9454545454545
348.205 19.2727272727273
349.205 19.0545454545455
350.205 19.2818181818182
351.205 18.6181818181818
352.205 18.5909090909091
353.205 18.4363636363636
354.205 18.7
355.205 18.1
356.205 18.1909090909091
357.205 20.3272727272727
358.205 21.9909090909091
359.205 24.3272727272727
360.205 26.2909090909091
361.205 27.4272727272727
362.205 29.4454545454545
363.205 31.2363636363636
364.205 33.1818181818182
365.205 34.6
366.205 37.2727272727273
367.205 38.2454545454545
368.205 38.2636363636364
369.205 38.7454545454545
370.205 39.7636363636364
371.205 40.6090909090909
372.205 42.2818181818182
373.205 42.9454545454545
374.205 43.2818181818182
375.205 44.5909090909091
376.205 45.4636363636364
377.205 43.7363636363636
378.205 45.1545454545455
379.205 45.2909090909091
380.205 45.8545454545455
381.205 45.8636363636364
382.205 45.3636363636364
383.205 42.3181818181818
384.205 42.4909090909091
385.205 41.5454545454545
386.205 39.6636363636364
387.205 37.7363636363636
388.205 38.8
389.205 36.7181818181818
390.205 35.0909090909091
391.205 33.0454545454545
392.205 31.0818181818182
393.205 29.1727272727273
394.205 30.4181818181818
395.205 28.5545454545455
396.205 28.2909090909091
397.205 28.3727272727273
398.205 28.5545454545455
399.205 29.7454545454545
400.205 31.4181818181818
401.205 33.2818181818182
402.205 34.9545454545454
403.205 36.1909090909091
404.205 38.3727272727273
405.205 39.9636363636364
406.205 41.3818181818182
407.205 43.2090909090909
408.205 44.6484848484848
409.205 46.9212121212121
410.205 47.0030303030303
411.205 46.6121212121212
412.205 44.4939393939394
415.205 42.3666666666667
416.205 39.9939393939394
417.205 37.8939393939394
418.205 35.5484848484848
419.205 33.4484848484849
420.205 30.8939393939394
421.205 29.1272727272727
422.205 26.4181818181818
423.205 23.7727272727273
424.205 22.2636363636364
425.205 22.2363636363636
426.205 23.2363636363636
427.205 24.4727272727273
428.205 25.8
429.205 26.6181818181818
430.205 27.9363636363636
431.205 29.5818181818182
432.205 30.7636363636364
433.205 31.9363636363636
434.205 33.7545454545455
435.205 34.8181818181818
436.205 35.8727272727273
437.205 36.5454545454545
438.205 36.7727272727273
439.205 36.7545454545455
440.205 37.3
441.205 36.8909090909091
442.205 36.6727272727273
443.205 36.4909090909091
444.205 37.1272727272727
445.205 35.9045454545455
446.205 36.2136363636364
447.205 36.1318181818182
448.205 33.9954545454545
449.205 34.2954545454546
451.205 34.6681818181818
452.205 34.5409090909091
453.205 35.3590909090909
454.205 35.7409090909091
455.205 35.5227272727273
456.205 35.9681818181818
457.205 37.3818181818182
458.205 37.5727272727273
459.205 37.5818181818182
460.205 38.6363636363636
461.205 36.8636363636364
462.205 36.4818181818182
463.205 36.2272727272727
464.205 35.1
465.205 34.4454545454545
466.205 34.4363636363636
467.205 33.2636363636364
468.205 32.5727272727273
469.205 31.7090909090909
470.205 31.9272727272727
471.205 31.9
472.205 33.2454545454545
473.205 32.8727272727273
474.205 32.4727272727273
475.205 32.8454545454545
476.205 32.1181818181818
477.205 31.3272727272727
478.205 31.9181818181818
479.205 31.9909090909091
480.205 32.2818181818182
481.205 32.3636363636364
482.205 32.5363636363636
483.205 32.5545454545455
484.205 32.9090909090909
485.205 33.8272727272727
486.205 33.6272727272727
487.205 34.5363636363636
488.205 35.9909090909091
489.205 36.0272727272727
490.205 36.5545454545455
491.205 36.6818181818182
492.205 37.0090909090909
493.205 38.3181818181818
494.205 37
495.205 34.7636363636364
496.205 34.8909090909091
497.205 35.9727272727273
498.205 36.6090909090909
499.205 36.6090909090909
500.205 37.1090909090909
501.205 37.5636363636364
502.205 37.8
503.205 35.9363636363636
504.205 35.9727272727273
505.205 38.3363636363636
506.205 40.9909090909091
507.205 41.3272727272727
508.205 40.6272727272727
509.205 40.9454545454545
510.205 41.1
511.205 40.8090909090909
512.205 40.5272727272727
513.205 40.5
514.205 42.9818181818182
515.205 42.3181818181818
516.205 42.7727272727273
517.205 42.9818181818182
518.205 42.7363636363636
519.205 43.4545454545455
520.205 42.8090909090909
521.205 42.8272727272727
522.205 43.5818181818182
523.205 43.0181818181818
524.205 43.2454545454545
525.205 43.0454545454545
526.205 43.2272727272727
527.205 42.5363636363636
528.205 42.5
529.205 42.3272727272727
530.205 42.4090909090909
531.205 43.2545454545455
532.205 43.6090909090909
533.205 43.4181818181818
534.205 44.2545454545455
535.205 42.0818181818182
536.205 42.0272727272727
537.205 42.5818181818182
538.205 42.6818181818182
539.205 42.5636363636364
540.205 43.4
541.205 43.3636363636364
542.205 43.4545454545455
543.205 42.9727272727273
544.205 43.0909090909091
545.205 42.6363636363636
546.205 44.9272727272727
547.205 45.2181818181818
548.205 44.8545454545455
549.205 45.0818181818182
550.205 45.4181818181818
551.205 44.8272727272727
552.205 44.8272727272727
553.205 44.5363636363636
554.205 44.5272727272727
555.205 43.0272727272727
556.205 42.4909090909091
557.205 42.9363636363636
558.205 43.2272727272727
559.205 44.2818181818182
560.205 45.0818181818182
561.205 45.7045454545455
562.205 46.6409090909091
563.205 47.4863636363636
564.205 47.6954545454545
565.205 48.6136363636364
567.205 48.6136363636364
568.205 49.7590909090909
569.205 49.8227272727273
570.205 49.7409090909091
571.205 48.5590909090909
572.205 48.3227272727273
573.205 47.6818181818182
574.205 47.3909090909091
575.205 46.7818181818182
576.205 46.5818181818182
577.205 45.1787878787879
578.205 43.6242424242424
579.205 43.9787878787879
580.205 44.1151515151515
581.205 44.0424242424242
584.205 44.8424242424242
585.205 44.4878787878788
586.205 44.5878787878788
587.205 44.2651515151515
588.205 44.6469696969697
589.205 44.4106060606061
590.205 45.5772727272727
591.205 47.4863636363636
593.205 46.9681818181818
594.205 46.6227272727273
595.205 44.0227272727273
596.205 43.3772727272727
597.205 43.0045454545455
598.205 43.15
599.205 43.3090909090909
600.205 43.1181818181818
601.205 43.7090909090909
602.205 44.1636363636364
603.205 45.6545454545455
605.205 45.7090909090909
606.205 45.4
607.205 48.1272727272727
608.205 48.7727272727273
609.205 49.0727272727273
610.205 48.7909090909091
611.205 49
612.205 48.7909090909091
613.205 48.7909090909091
614.205 48.4318181818182
615.205 48.1545454545455
617.205 48.2909090909091
618.205 47.2909090909091
619.205 46.5272727272727
621.205 43.1169696969697
624.205 39.649696969697
625.205 36.6315151515152
626.205 33.1075757575758
629.205 29.8712121212121
630.205 26.5348484848485
631.205 22.830303030303
634.205 19.6439393939394
635.205 16.3893939393939
636.205 14.2348484848485
637.205 11.4893939393939
638.205 11.3906060606061
639.205 11.5851515151515
640.205 11.4578787878788
641.205 11.3404545454545
642.205 11.0313636363636
643.205 10.6713636363636
644.205 10.4540909090909
646.205 10.0295454545455
647.205 9.76409090909091
648.205 9.74681818181818
649.205 9.53681818181818
650.205 9.115
651.205 8.88590909090909
652.205 8.70409090909091
653.205 8.57272727272727
654.205 8.39272727272727
655.205 8.44363636363637
656.205 8.67
657.205 8.41363636363636
658.205 8.38
659.205 8.22636363636364
660.205 8.10636363636364
661.205 8.26909090909091
662.205 8.17636363636364
663.205 7.95727272727273
664.205 7.75181818181818
665.205 7.92272727272727
666.205 7.72636363636364
667.205 7.50909090909091
668.205 7.8
669.205 7.85090909090909
670.205 7.72272727272727
671.205 7.56090909090909
672.205 7.58636363636364
673.205 7.53363636363636
674.205 7.56
675.205 7.89363636363636
676.205 7.65454545454546
677.205 7.72272727272727
678.205 7.57727272727273
679.205 7.44909090909091
680.205 7.28727272727273
681.205 7.39
682.205 7.48363636363636
683.205 7.36363636363636
684.205 7.32090909090909
685.205 7.30454545454545
686.205 6.95454545454545
687.205 7.16
688.205 6.83545454545455
689.205 6.87
690.205 6.69
691.205 6.67272727272727
692.205 6.65545454545455
693.205 6.50181818181818
694.205 6.11727272727273
695.205 5.96363636363636
696.205 5.62090909090909
697.205 5.44090909090909
698.205 5.37090909090909
699.205 5.16636363636364
700.205 5.00363636363636
701.205 4.96090909090909
702.205 4.79
703.205 4.82272727272727
704.205 4.56727272727273
705.205 4.63545454545454
706.205 4.60090909090909
707.205 4.51545454545455
708.205 4.60090909090909
709.205 4.47272727272727
710.205 4.66909090909091
711.205 4.84
712.205 4.72
713.205 4.88181818181818
714.205 4.84090909090909
715.205 5.07
716.205 5.18181818181818
717.205 5.41272727272727
718.205 5.72090909090909
719.205 6.13090909090909
720.205 6.10727272727273
721.205 6.08090909090909
722.205 5.97818181818182
723.205 6.13272727272727
724.205 6.23545454545455
725.205 6.39727272727273
726.205 6.48363636363636
727.205 6.64545454545454
728.205 6.73090909090909
729.205 6.85909090909091
730.205 6.44090909090909
731.205 6.32818181818182
732.205 6.51636363636364
733.205 6.15
734.205 5.95181818181818
735.205 5.65363636363636
736.205 5.28636363636364
737.205 5.04636363636364
738.205 4.80727272727273
739.205 4.49909090909091
740.205 4.04545454545455
741.205 4.13090909090909
742.205 4.06272727272727
743.205 4.05272727272727
744.205 4.20636363636364
745.205 4.26636363636364
746.205 4.16363636363636
747.205 4.09545454545455
748.205 4.01818181818182
749.205 3.88090909090909
750.205 3.76181818181818
751.205 3.87363636363636
752.205 3.73636363636364
753.205 3.78
754.205 3.55090909090909
755.205 3.51545454545455
756.205 3.57636363636364
757.205 3.50727272727273
758.205 3.72
759.205 3.55909090909091
760.205 3.70454545454545
761.205 3.71272727272727
762.205 3.73818181818182
763.205 3.78090909090909
764.205 3.68636363636364
765.205 3.71181818181818
766.205 3.89909090909091
767.205 3.82363636363636
768.205 3.95909090909091
769.205 3.57454545454546
770.205 3.81363636363636
771.205 3.97636363636364
772.205 3.84
773.205 3.66
774.205 3.71909090909091
775.205 3.74545454545455
776.205 3.81272727272727
777.205 3.49727272727273
778.205 3.36818181818182
779.205 3.36
780.205 3.34363636363636
781.205 3.11181818181818
782.205 2.83
783.205 3.03454545454545
784.205 2.99787878787879
785.205 2.65787878787879
786.205 2.68151515151515
787.205 2.4869696969697
788.205 2.64878787878788
791.205 2.52060606060606
792.205 2.4269696969697
793.205 2.50242424242424
794.205 2.63060606060606
795.205 2.61333333333333
796.205 2.4169696969697
797.205 2.47090909090909
798.205 2.70090909090909
799.205 2.67727272727273
800.205 2.59090909090909
801.205 2.51454545454545
802.205 2.65818181818182
803.205 2.59
805.205 2.60909090909091
806.205 2.61636363636364
807.205 2.45454545454545
808.205 2.58272727272727
809.205 2.62454545454545
810.205 2.79545454545455
811.205 2.66636363636364
812.205 2.74363636363636
813.205 2.80363636363636
814.205 2.79454545454545
816.205 2.75181818181818
817.205 2.77636363636364
818.205 2.70909090909091
819.205 2.75666666666667
820.205 2.73939393939394
821.205 2.65939393939394
822.205 2.2830303030303
823.205 2.64212121212121
826.205 2.5730303030303
827.205 2.70848484848485
829.205 2.7430303030303
830.205 2.75212121212121
831.205 2.5730303030303
833.205 2.6330303030303
835.205 2.71272727272727
836.205 2.58545454545455
837.205 2.71727272727273
838.205 2.87909090909091
839.205 2.53
840.205 2.77727272727273
841.205 2.64909090909091
843.205 2.63272727272727
844.205 2.80454545454545
845.205 2.96636363636364
847.205 2.94909090909091
848.205 3
849.205 3.03363636363636
850.205 2.94
852.205 2.96636363636364
853.205 2.95636363636364
854.205 2.79454545454545
855.205 2.77
856.205 2.61636363636364
857.205 2.55636363636364
858.205 2.64090909090909
859.205 2.53
860.205 2.41454545454545
861.205 2.5
862.205 2.37272727272727
864.205 2.33727272727273
865.205 2.33818181818182
867.205 2.24454545454545
868.205 2.31363636363636
869.205 2.47545454545455
870.205 2.35454545454545
871.205 2.11727272727273
872.205 2.33090909090909
873.205 2.26727272727273
874.205 2.29181818181818
875.205 2.29090909090909
876.205 2.20818181818182
877.205 2.35181818181818
878.205 2.63363636363636
879.205 2.64181818181818
880.205 2.60727272727273
881.205 2.60818181818182
882.205 2.74454545454545
883.205 2.65818181818182
885.205 2.65090909090909
886.205 2.60909090909091
887.205 2.71909090909091
888.205 2.79454545454545
889.205 2.77727272727273
890.205 2.79363636363636
891.205 2.89545454545455
892.205 2.76818181818182
893.205 2.903
895.205 2.90111111111111
897.205 2.9125
899.205 3.13857142857143
901.205 3.23833333333333
};

\addplot [very thick, red, const plot mark left, dotted]
table {%
0 100
13.096 100
328.543 100
341.84 100
356.642 100
372.244 100
387.817 100
400.945 100
416.102 100
429.233 100
443.406 100
461.871 100
476.016 100
492.186 100
505.57 100
517.477 100
533.861 100
559.119 100
622.102 10
640.937 8
659.593 7
672.893 7
687.853 6
702.811 6
716.316 10
730.48 8
744.879 6
759.053 8
772.224 10
787.438 15
802.795 25
816.748 20
830.86 30
844.847 40
865.347 10
879.599 20
893.551 10
};

\addplot [purple, opacity=0.7, densely dashdotted, forget plot, very thick]
table {%
308.288 0
308.288 104.894136363636
};
\addplot [purple, opacity=0.7,  densely dashdotted, forget plot, very thick]
table {%
605.356 0
605.356 104.894136363636
};




\end{axis}

\begin{axis}[
width=\fwidth,
height=\fheight,
axis y line*=right,
tick align=outside,
x grid style={darkgrey176},
xtick=\empty,  
xmin=200, xmax=901.933,
xtick pos=left,
xtick style={color=black},
y grid style={darkgrey176},
ylabel=\textcolor{green}{Target SNR [dB]},
ymin=0, ymax=19,
ytick pos=right,
ytick style={color=green},
yticklabel style={anchor=west}, 
ylabel style={font=\footnotesize\color{white!15!black}},
ylabel shift=-4pt
]
\addplot [very thick, green01270, const plot mark left, dashed]
table {%
0 18
4.515 18
22.714 18
43.618 18
63.878 18
80.862 18
100.344 18
118.337 18
137.375 18
155.581 18
176.273 18
196.899 18
214.707 18
235.567 18
258.666 18
282.159 18
311.749 18
333.864 18
356.027 18
372.666 18
390.696 18
408.731 18
439.918 18
456.946 18
475.402 18
494.895 18
511.521 18
530.393 18
551.318 18
572.596 18
597.562 18
618.436 18
640.732 15
661.426 12
679.64 9
696.448 6
711.812 9
730.685 6
746.725 3
763.173 3
779.416 0
795.415 0
810.78 0
828.206 0
844.429 0
862.698 0
880.008 0
897.65 0
};

\end{axis}

\end{tikzpicture}

%% file: figures/power_consumption.tex
\begin{tikzpicture}
\pgfplotsset{every tick label/.append style={font=\scriptsize}}

\definecolor{darkgrey176}{RGB}{176,176,176}
\definecolor{green}{RGB}{0,128,0}
\definecolor{green01270}{RGB}{0,127,0}
\definecolor{lightgrey204}{RGB}{204,204,204}
\definecolor{purple}{RGB}{128,0,128}

\begin{axis}[
legend cell align={left},
legend style={fill opacity=0.8, draw opacity=1, text opacity=1, draw=lightgrey204},
tick align=outside,
tick pos=left,
x grid style={darkgrey176},
xlabel={Time [s]},
xmajorgrids,
xmin=200, xmax=901.933,
xtick style={color=black},
y grid style={darkgrey176},
ylabel={Power [W]},
ymajorgrids,
ymin=0.93195, ymax=2.05103181818182,
ytick style={color=black},
width=\fwidth,
height=\fheight,
xlabel style={font=\footnotesize\color{white!15!black}},
ylabel style={font=\footnotesize\color{white!15!black}},
xlabel shift=-2pt,
ylabel shift=-5pt
]

\fill[red!10, opacity=0.5] (axis cs:308.288,0) rectangle (axis cs:605.356,2.06);

\path [draw=green, fill=green, opacity=0.3]
(axis cs:5.942,0)
--(axis cs:5.942,0.981)
--(axis cs:7.37,1.02028571428571)
--(axis cs:10.027,1.096375)
--(axis cs:12.674,1.20455555555556)
--(axis cs:14.13,1.2965)
--(axis cs:15.353,1.36690909090909)
--(axis cs:18.008,1.45672727272727)
--(axis cs:20.667,1.53263636363636)
--(axis cs:22.098,1.59918181818182)
--(axis cs:22.917,1.63609090909091)
--(axis cs:24.141,1.655)
--(axis cs:24.967,1.66009090909091)
--(axis cs:26.194,1.63090909090909)
--(axis cs:27.022,1.56945454545455)
--(axis cs:27.839,1.46872727272727)
--(axis cs:28.669,1.36481818181818)
--(axis cs:29.53,1.266)
--(axis cs:30.347,1.17754545454545)
--(axis cs:31.165,1.10681818181818)
--(axis cs:32.004,1.05472727272727)
--(axis cs:32.973,1.071)
--(axis cs:33.791,1.11345454545455)
--(axis cs:34.609,1.15872727272727)
--(axis cs:35.434,1.22454545454545)
--(axis cs:36.458,1.30336363636364)
--(axis cs:39.935,1.35936363636364)
--(axis cs:41.568,1.41272727272727)
--(axis cs:56.712,1.466)
--(axis cs:72.459,1.50881818181818)
--(axis cs:73.303,1.57636363636364)
--(axis cs:74.119,1.60954545454545)
--(axis cs:75.137,1.60763636363636)
--(axis cs:75.952,1.595)
--(axis cs:82.699,1.59463636363636)
--(axis cs:83.727,1.58118181818182)
--(axis cs:99.934,1.55327272727273)
--(axis cs:103.21,1.54345454545455)
--(axis cs:104.029,1.53545454545455)
--(axis cs:108.937,1.52690909090909)
--(axis cs:122.419,1.52927272727273)
--(axis cs:149.867,1.50627272727273)
--(axis cs:153.136,1.51072727272727)
--(axis cs:153.952,1.50763636363636)
--(axis cs:154.766,1.50827272727273)
--(axis cs:167.684,1.50463636363636)
--(axis cs:171.976,1.50509090909091)
--(axis cs:178.525,1.50027272727273)
--(axis cs:179.54,1.50709090909091)
--(axis cs:186.91,1.51545454545455)
--(axis cs:188.545,1.51845454545455)
--(axis cs:194.044,1.52363636363636)
--(axis cs:203.039,1.52309090909091)
--(axis cs:206.324,1.52236363636364)
--(axis cs:210.416,1.52081818181818)
--(axis cs:211.232,1.52163636363636)
--(axis cs:217.784,1.52145454545455)
--(axis cs:218.598,1.51872727272727)
--(axis cs:222.077,1.52554545454545)
--(axis cs:226.18,1.52036363636364)
--(axis cs:233.739,1.514)
--(axis cs:242.306,1.52127272727273)
--(axis cs:248.255,1.528)
--(axis cs:249.083,1.52563636363636)
--(axis cs:249.897,1.52763636363636)
--(axis cs:251.319,1.52527272727273)
--(axis cs:256.009,1.52454545454545)
--(axis cs:258.671,1.52909090909091)
--(axis cs:265.812,1.53209090909091)
--(axis cs:267.648,1.52863636363636)
--(axis cs:283.589,1.52963636363636)
--(axis cs:286.877,1.53109090909091)
--(axis cs:287.695,1.52572727272727)
--(axis cs:291.77,1.51518181818182)
--(axis cs:300.141,1.51854545454545)
--(axis cs:306.246,1.51936363636364)
--(axis cs:316.047,1.523)
--(axis cs:317.886,1.52790909090909)
--(axis cs:321.175,1.52409090909091)
--(axis cs:326.292,1.52545454545455)
--(axis cs:327.727,1.53063636363636)
--(axis cs:328.75,1.53054545454545)
--(axis cs:330.378,1.53490909090909)
--(axis cs:338.774,1.54236363636364)
--(axis cs:340.402,1.54463636363636)
--(axis cs:343.48,1.54736363636364)
--(axis cs:344.498,1.54845454545455)
--(axis cs:346.145,1.54981818181818)
--(axis cs:347.799,1.54754545454545)
--(axis cs:351.073,1.54736363636364)
--(axis cs:353.739,1.54309090909091)
--(axis cs:354.559,1.53881818181818)
--(axis cs:355.401,1.55218181818182)
--(axis cs:356.233,1.55627272727273)
--(axis cs:357.051,1.56090909090909)
--(axis cs:358.678,1.57090909090909)
--(axis cs:359.495,1.58690909090909)
--(axis cs:361.967,1.58254545454545)
--(axis cs:362.789,1.59872727272727)
--(axis cs:363.619,1.61472727272727)
--(axis cs:366.08,1.61336363636364)
--(axis cs:366.907,1.59763636363636)
--(axis cs:369.355,1.60818181818182)
--(axis cs:370.174,1.60718181818182)
--(axis cs:374.3,1.60136363636364)
--(axis cs:376.348,1.59418181818182)
--(axis cs:381.891,1.58409090909091)
--(axis cs:385.16,1.57672727272727)
--(axis cs:391.721,1.59063636363636)
--(axis cs:395.2,1.57436363636364)
--(axis cs:396.846,1.56072727272727)
--(axis cs:400.121,1.56927272727273)
--(axis cs:400.95,1.59563636363636)
--(axis cs:407.705,1.59536363636364)
--(axis cs:409.344,1.59354545454545)
--(axis cs:410.162,1.59345454545455)
--(axis cs:411.811,1.59027272727273)
--(axis cs:412.627,1.59354545454545)
--(axis cs:416.308,1.58236363636364)
--(axis cs:417.137,1.56945454545455)
--(axis cs:420.425,1.57036363636364)
--(axis cs:421.244,1.56954545454545)
--(axis cs:422.884,1.55327272727273)
--(axis cs:426.986,1.55936363636364)
--(axis cs:428.617,1.56754545454545)
--(axis cs:429.453,1.56918181818182)
--(axis cs:430.282,1.56690909090909)
--(axis cs:431.1,1.57654545454545)
--(axis cs:432.751,1.58409090909091)
--(axis cs:433.57,1.58336363636364)
--(axis cs:434.394,1.58954545454545)
--(axis cs:436.029,1.61663636363636)
--(axis cs:441.759,1.61145454545455)
--(axis cs:442.59,1.63290909090909)
--(axis cs:444.43,1.63618181818182)
--(axis cs:446.905,1.63045454545455)
--(axis cs:451.414,1.63336363636364)
--(axis cs:454.281,1.64618181818182)
--(axis cs:456.951,1.64818181818182)
--(axis cs:457.767,1.64327272727273)
--(axis cs:458.598,1.66927272727273)
--(axis cs:462.693,1.686)
--(axis cs:464.337,1.68109090909091)
--(axis cs:465.171,1.68781818181818)
--(axis cs:466.813,1.64881818181818)
--(axis cs:467.628,1.587)
--(axis cs:468.447,1.52463636363636)
--(axis cs:470.079,1.46718181818182)
--(axis cs:470.897,1.40690909090909)
--(axis cs:472.53,1.34790909090909)
--(axis cs:473.341,1.30954545454545)
--(axis cs:474.151,1.28554545454545)
--(axis cs:474.982,1.30272727272727)
--(axis cs:475.809,1.31945454545455)
--(axis cs:476.628,1.35227272727273)
--(axis cs:477.456,1.43536363636364)
--(axis cs:479.096,1.53254545454545)
--(axis cs:479.912,1.64990909090909)
--(axis cs:480.725,1.73436363636364)
--(axis cs:481.558,1.81763636363636)
--(axis cs:482.379,1.88409090909091)
--(axis cs:483.202,1.92563636363636)
--(axis cs:484.021,1.94009090909091)
--(axis cs:484.837,1.90736363636364)
--(axis cs:486.069,1.87845454545455)
--(axis cs:488.102,1.84345454545455)
--(axis cs:488.918,1.79945454545455)
--(axis cs:489.944,1.76254545454545)
--(axis cs:491.775,1.70063636363636)
--(axis cs:492.595,1.67745454545455)
--(axis cs:493.413,1.62481818181818)
--(axis cs:494.224,1.59136363636364)
--(axis cs:495.101,1.58436363636364)
--(axis cs:495.934,1.58527272727273)
--(axis cs:498.398,1.59627272727273)
--(axis cs:499.216,1.60681818181818)
--(axis cs:500.03,1.62254545454545)
--(axis cs:500.843,1.61536363636364)
--(axis cs:501.678,1.59672727272727)
--(axis cs:503.323,1.56245454545455)
--(axis cs:504.141,1.49709090909091)
--(axis cs:504.957,1.466)
--(axis cs:505.78,1.43409090909091)
--(axis cs:506.595,1.46490909090909)
--(axis cs:507.411,1.51618181818182)
--(axis cs:508.243,1.55609090909091)
--(axis cs:509.067,1.59154545454545)
--(axis cs:509.882,1.60590909090909)
--(axis cs:511.524,1.61836363636364)
--(axis cs:512.342,1.629)
--(axis cs:513.163,1.66554545454545)
--(axis cs:513.978,1.72463636363636)
--(axis cs:514.806,1.77818181818182)
--(axis cs:516.44,1.83154545454545)
--(axis cs:517.257,1.79754545454545)
--(axis cs:518.095,1.73781818181818)
--(axis cs:519.731,1.68981818181818)
--(axis cs:520.774,1.64309090909091)
--(axis cs:523.229,1.62118181818182)
--(axis cs:525.474,1.61318181818182)
--(axis cs:527.119,1.61309090909091)
--(axis cs:527.932,1.61745454545455)
--(axis cs:528.751,1.62518181818182)
--(axis cs:529.576,1.62754545454545)
--(axis cs:530.395,1.63372727272727)
--(axis cs:531.21,1.64363636363636)
--(axis cs:532.026,1.64427272727273)
--(axis cs:535.521,1.63527272727273)
--(axis cs:537.167,1.62645454545455)
--(axis cs:537.986,1.62336363636364)
--(axis cs:538.809,1.61981818181818)
--(axis cs:539.627,1.62418181818182)
--(axis cs:540.441,1.62045454545455)
--(axis cs:541.261,1.60990909090909)
--(axis cs:542.1,1.61163636363636)
--(axis cs:543.727,1.60718181818182)
--(axis cs:544.548,1.607)
--(axis cs:545.376,1.61863636363636)
--(axis cs:547.825,1.62327272727273)
--(axis cs:548.658,1.62918181818182)
--(axis cs:549.475,1.62745454545455)
--(axis cs:550.292,1.62927272727273)
--(axis cs:551.112,1.63118181818182)
--(axis cs:551.938,1.64027272727273)
--(axis cs:552.753,1.629)
--(axis cs:553.572,1.62872727272727)
--(axis cs:554.399,1.60281818181818)
--(axis cs:555.421,1.61254545454545)
--(axis cs:556.239,1.61081818181818)
--(axis cs:557.462,1.63109090909091)
--(axis cs:559.123,1.63972727272727)
--(axis cs:560.548,1.66090909090909)
--(axis cs:561.365,1.66590909090909)
--(axis cs:563.194,1.66345454545455)
--(axis cs:564.444,1.65)
--(axis cs:565.259,1.66218181818182)
--(axis cs:568.715,1.66109090909091)
--(axis cs:570.558,1.632)
--(axis cs:572.185,1.56263636363636)
--(axis cs:573.014,1.49845454545455)
--(axis cs:574.845,1.44627272727273)
--(axis cs:575.687,1.452)
--(axis cs:578.744,1.47227272727273)
--(axis cs:579.577,1.506)
--(axis cs:581.611,1.55081818181818)
--(axis cs:584.474,1.61263636363636)
--(axis cs:585.301,1.66727272727273)
--(axis cs:586.125,1.71872727272727)
--(axis cs:587.353,1.79772727272727)
--(axis cs:588.171,1.86081818181818)
--(axis cs:588.995,1.92127272727273)
--(axis cs:590.421,1.95336363636364)
--(axis cs:591.237,1.93690909090909)
--(axis cs:594.286,1.89154545454545)
--(axis cs:595.715,1.85)
--(axis cs:596.94,1.81363636363636)
--(axis cs:598.583,1.75054545454545)
--(axis cs:599.405,1.70536363636364)
--(axis cs:601.049,1.66663636363636)
--(axis cs:601.867,1.65)
--(axis cs:603.296,1.64818181818182)
--(axis cs:605.359,1.649)
--(axis cs:607.389,1.65972727272727)
--(axis cs:608.206,1.67118181818182)
--(axis cs:609.031,1.69209090909091)
--(axis cs:609.85,1.64618181818182)
--(axis cs:610.667,1.66772727272727)
--(axis cs:611.892,1.64227272727273)
--(axis cs:612.717,1.61872727272727)
--(axis cs:614.351,1.57627272727273)
--(axis cs:618.019,1.54427272727273)
--(axis cs:619.863,1.51263636363636)
--(axis cs:630.882,1.48109090909091)
--(axis cs:636.856,1.44418181818182)
--(axis cs:640.939,1.39609090909091)
--(axis cs:641.754,1.34527272727273)
--(axis cs:646.651,1.36590909090909)
--(axis cs:652.2,1.30954545454545)
--(axis cs:658.97,1.31327272727273)
--(axis cs:662.253,1.30318181818182)
--(axis cs:666.754,1.30872727272727)
--(axis cs:667.773,1.31781818181818)
--(axis cs:670.232,1.33036363636364)
--(axis cs:673.712,1.32236363636364)
--(axis cs:674.528,1.31436363636364)
--(axis cs:681.483,1.32663636363636)
--(axis cs:687.44,1.32718181818182)
--(axis cs:689.893,1.33063636363636)
--(axis cs:690.716,1.32645454545455)
--(axis cs:691.556,1.33018181818182)
--(axis cs:692.371,1.33463636363636)
--(axis cs:693.999,1.35327272727273)
--(axis cs:694.82,1.35236363636364)
--(axis cs:696.45,1.353)
--(axis cs:698.083,1.37345454545455)
--(axis cs:698.899,1.38181818181818)
--(axis cs:699.731,1.39427272727273)
--(axis cs:703.019,1.41818181818182)
--(axis cs:707.937,1.42954545454545)
--(axis cs:710.385,1.44045454545455)
--(axis cs:714.06,1.44990909090909)
--(axis cs:716.318,1.44381818181818)
--(axis cs:719.604,1.43209090909091)
--(axis cs:721.656,1.42781818181818)
--(axis cs:724.934,1.42527272727273)
--(axis cs:725.762,1.42281818181818)
--(axis cs:734.798,1.432)
--(axis cs:736.452,1.43054545454545)
--(axis cs:739.74,1.42718181818182)
--(axis cs:743.858,1.41545454545455)
--(axis cs:747.137,1.417)
--(axis cs:747.954,1.41445454545455)
--(axis cs:748.788,1.42209090909091)
--(axis cs:749.613,1.42618181818182)
--(axis cs:752.684,1.43127272727273)
--(axis cs:754.764,1.43)
--(axis cs:756.189,1.42281818181818)
--(axis cs:761.101,1.41845454545455)
--(axis cs:763.584,1.409)
--(axis cs:764.4,1.40836363636364)
--(axis cs:766.874,1.41718181818182)
--(axis cs:767.689,1.40772727272727)
--(axis cs:770.17,1.40354545454545)
--(axis cs:774.27,1.40145454545455)
--(axis cs:775.09,1.40036363636364)
--(axis cs:776.755,1.39672727272727)
--(axis cs:777.575,1.39436363636364)
--(axis cs:778.39,1.40218181818182)
--(axis cs:780.036,1.40436363636364)
--(axis cs:781.676,1.40918181818182)
--(axis cs:784.962,1.40681818181818)
--(axis cs:792.564,1.40654545454545)
--(axis cs:794.602,1.41309090909091)
--(axis cs:798.093,1.41190909090909)
--(axis cs:804.621,1.41509090909091)
--(axis cs:807.919,1.41381818181818)
--(axis cs:809.553,1.41390909090909)
--(axis cs:810.367,1.41590909090909)
--(axis cs:812.001,1.41081818181818)
--(axis cs:813.627,1.40918181818182)
--(axis cs:816.96,1.41436363636364)
--(axis cs:817.775,1.409)
--(axis cs:818.59,1.40590909090909)
--(axis cs:819.404,1.39481818181818)
--(axis cs:820.236,1.40327272727273)
--(axis cs:822.685,1.39418181818182)
--(axis cs:823.51,1.39481818181818)
--(axis cs:824.329,1.39281818181818)
--(axis cs:826.771,1.38909090909091)
--(axis cs:827.594,1.39072727272727)
--(axis cs:828.822,1.38790909090909)
--(axis cs:830.044,1.381)
--(axis cs:831.274,1.38190909090909)
--(axis cs:832.297,1.38018181818182)
--(axis cs:833.132,1.38927272727273)
--(axis cs:833.959,1.38272727272727)
--(axis cs:834.787,1.38518181818182)
--(axis cs:835.615,1.37790909090909)
--(axis cs:837.26,1.37454545454545)
--(axis cs:838.081,1.38036363636364)
--(axis cs:838.899,1.38181818181818)
--(axis cs:841.357,1.385)
--(axis cs:842.176,1.40190909090909)
--(axis cs:843.813,1.40854545454545)
--(axis cs:848.121,1.41381818181818)
--(axis cs:848.938,1.41672727272727)
--(axis cs:851.417,1.42409090909091)
--(axis cs:852.245,1.418)
--(axis cs:854.714,1.42036363636364)
--(axis cs:855.529,1.42290909090909)
--(axis cs:857.171,1.41263636363636)
--(axis cs:857.99,1.40936363636364)
--(axis cs:858.806,1.41245454545455)
--(axis cs:860.443,1.40218181818182)
--(axis cs:863.312,1.406)
--(axis cs:864.731,1.40518181818182)
--(axis cs:865.554,1.40545454545455)
--(axis cs:868.222,1.39818181818182)
--(axis cs:869.056,1.39490909090909)
--(axis cs:870.081,1.39272727272727)
--(axis cs:871.736,1.39309090909091)
--(axis cs:873.379,1.40554545454545)
--(axis cs:876.094,1.392)
--(axis cs:877.749,1.39481818181818)
--(axis cs:878.567,1.38945454545455)
--(axis cs:879.393,1.38436363636364)
--(axis cs:880.214,1.38381818181818)
--(axis cs:881.029,1.38654545454545)
--(axis cs:881.844,1.38445454545455)
--(axis cs:882.665,1.39581818181818)
--(axis cs:883.486,1.39890909090909)
--(axis cs:884.299,1.395)
--(axis cs:885.109,1.37181818181818)
--(axis cs:885.99,1.38727272727273)
--(axis cs:886.804,1.37263636363636)
--(axis cs:887.618,1.37218181818182)
--(axis cs:888.434,1.36827272727273)
--(axis cs:889.258,1.35381818181818)
--(axis cs:890.075,1.34581818181818)
--(axis cs:891.712,1.35036363636364)
--(axis cs:892.528,1.34909090909091)
--(axis cs:893.345,1.34681818181818)
--(axis cs:894.168,1.35)
--(axis cs:894.982,1.369)
--(axis cs:895.795,1.37181818181818)
--(axis cs:896.626,1.38318181818182)
--(axis cs:897.444,1.3828)
--(axis cs:898.26,1.38377777777778)
--(axis cs:899.08,1.398375)
--(axis cs:899.894,1.40285714285714)
--(axis cs:901.933,1.40183333333333)
--(axis cs:901.933,0)
--(axis cs:901.933,0)
--(axis cs:899.894,0)
--(axis cs:899.08,0)
--(axis cs:898.26,0)
--(axis cs:897.444,0)
--(axis cs:896.626,0)
--(axis cs:895.795,0)
--(axis cs:894.982,0)
--(axis cs:894.168,0)
--(axis cs:893.345,0)
--(axis cs:892.528,0)
--(axis cs:891.712,0)
--(axis cs:890.075,0)
--(axis cs:889.258,0)
--(axis cs:888.434,0)
--(axis cs:887.618,0)
--(axis cs:886.804,0)
--(axis cs:885.99,0)
--(axis cs:885.109,0)
--(axis cs:884.299,0)
--(axis cs:883.486,0)
--(axis cs:882.665,0)
--(axis cs:881.844,0)
--(axis cs:881.029,0)
--(axis cs:880.214,0)
--(axis cs:879.393,0)
--(axis cs:878.567,0)
--(axis cs:877.749,0)
--(axis cs:876.094,0)
--(axis cs:873.379,0)
--(axis cs:871.736,0)
--(axis cs:870.081,0)
--(axis cs:869.056,0)
--(axis cs:868.222,0)
--(axis cs:865.554,0)
--(axis cs:864.731,0)
--(axis cs:863.312,0)
--(axis cs:860.443,0)
--(axis cs:858.806,0)
--(axis cs:857.99,0)
--(axis cs:857.171,0)
--(axis cs:855.529,0)
--(axis cs:854.714,0)
--(axis cs:852.245,0)
--(axis cs:851.417,0)
--(axis cs:848.938,0)
--(axis cs:848.121,0)
--(axis cs:843.813,0)
--(axis cs:842.176,0)
--(axis cs:841.357,0)
--(axis cs:838.899,0)
--(axis cs:838.081,0)
--(axis cs:837.26,0)
--(axis cs:835.615,0)
--(axis cs:834.787,0)
--(axis cs:833.959,0)
--(axis cs:833.132,0)
--(axis cs:832.297,0)
--(axis cs:831.274,0)
--(axis cs:830.044,0)
--(axis cs:828.822,0)
--(axis cs:827.594,0)
--(axis cs:826.771,0)
--(axis cs:824.329,0)
--(axis cs:823.51,0)
--(axis cs:822.685,0)
--(axis cs:820.236,0)
--(axis cs:819.404,0)
--(axis cs:818.59,0)
--(axis cs:817.775,0)
--(axis cs:816.96,0)
--(axis cs:813.627,0)
--(axis cs:812.001,0)
--(axis cs:810.367,0)
--(axis cs:809.553,0)
--(axis cs:807.919,0)
--(axis cs:804.621,0)
--(axis cs:798.093,0)
--(axis cs:794.602,0)
--(axis cs:792.564,0)
--(axis cs:784.962,0)
--(axis cs:781.676,0)
--(axis cs:780.036,0)
--(axis cs:778.39,0)
--(axis cs:777.575,0)
--(axis cs:776.755,0)
--(axis cs:775.09,0)
--(axis cs:774.27,0)
--(axis cs:770.17,0)
--(axis cs:767.689,0)
--(axis cs:766.874,0)
--(axis cs:764.4,0)
--(axis cs:763.584,0)
--(axis cs:761.101,0)
--(axis cs:756.189,0)
--(axis cs:754.764,0)
--(axis cs:752.684,0)
--(axis cs:749.613,0)
--(axis cs:748.788,0)
--(axis cs:747.954,0)
--(axis cs:747.137,0)
--(axis cs:743.858,0)
--(axis cs:739.74,0)
--(axis cs:736.452,0)
--(axis cs:734.798,0)
--(axis cs:725.762,0)
--(axis cs:724.934,0)
--(axis cs:721.656,0)
--(axis cs:719.604,0)
--(axis cs:716.318,0)
--(axis cs:714.06,0)
--(axis cs:710.385,0)
--(axis cs:707.937,0)
--(axis cs:703.019,0)
--(axis cs:699.731,0)
--(axis cs:698.899,0)
--(axis cs:698.083,0)
--(axis cs:696.45,0)
--(axis cs:694.82,0)
--(axis cs:693.999,0)
--(axis cs:692.371,0)
--(axis cs:691.556,0)
--(axis cs:690.716,0)
--(axis cs:689.893,0)
--(axis cs:687.44,0)
--(axis cs:681.483,0)
--(axis cs:674.528,0)
--(axis cs:673.712,0)
--(axis cs:670.232,0)
--(axis cs:667.773,0)
--(axis cs:666.754,0)
--(axis cs:662.253,0)
--(axis cs:658.97,0)
--(axis cs:652.2,0)
--(axis cs:646.651,0)
--(axis cs:641.754,0)
--(axis cs:640.939,0)
--(axis cs:636.856,0)
--(axis cs:630.882,0)
--(axis cs:619.863,0)
--(axis cs:618.019,0)
--(axis cs:614.351,0)
--(axis cs:612.717,0)
--(axis cs:611.892,0)
--(axis cs:610.667,0)
--(axis cs:609.85,0)
--(axis cs:609.031,0)
--(axis cs:608.206,0)
--(axis cs:607.389,0)
--(axis cs:605.359,0)
--(axis cs:603.296,0)
--(axis cs:601.867,0)
--(axis cs:601.049,0)
--(axis cs:599.405,0)
--(axis cs:598.583,0)
--(axis cs:596.94,0)
--(axis cs:595.715,0)
--(axis cs:594.286,0)
--(axis cs:591.237,0)
--(axis cs:590.421,0)
--(axis cs:588.995,0)
--(axis cs:588.171,0)
--(axis cs:587.353,0)
--(axis cs:586.125,0)
--(axis cs:585.301,0)
--(axis cs:584.474,0)
--(axis cs:581.611,0)
--(axis cs:579.577,0)
--(axis cs:578.744,0)
--(axis cs:575.687,0)
--(axis cs:574.845,0)
--(axis cs:573.014,0)
--(axis cs:572.185,0)
--(axis cs:570.558,0)
--(axis cs:568.715,0)
--(axis cs:565.259,0)
--(axis cs:564.444,0)
--(axis cs:563.194,0)
--(axis cs:561.365,0)
--(axis cs:560.548,0)
--(axis cs:559.123,0)
--(axis cs:557.462,0)
--(axis cs:556.239,0)
--(axis cs:555.421,0)
--(axis cs:554.399,0)
--(axis cs:553.572,0)
--(axis cs:552.753,0)
--(axis cs:551.938,0)
--(axis cs:551.112,0)
--(axis cs:550.292,0)
--(axis cs:549.475,0)
--(axis cs:548.658,0)
--(axis cs:547.825,0)
--(axis cs:545.376,0)
--(axis cs:544.548,0)
--(axis cs:543.727,0)
--(axis cs:542.1,0)
--(axis cs:541.261,0)
--(axis cs:540.441,0)
--(axis cs:539.627,0)
--(axis cs:538.809,0)
--(axis cs:537.986,0)
--(axis cs:537.167,0)
--(axis cs:535.521,0)
--(axis cs:532.026,0)
--(axis cs:531.21,0)
--(axis cs:530.395,0)
--(axis cs:529.576,0)
--(axis cs:528.751,0)
--(axis cs:527.932,0)
--(axis cs:527.119,0)
--(axis cs:525.474,0)
--(axis cs:523.229,0)
--(axis cs:520.774,0)
--(axis cs:519.731,0)
--(axis cs:518.095,0)
--(axis cs:517.257,0)
--(axis cs:516.44,0)
--(axis cs:514.806,0)
--(axis cs:513.978,0)
--(axis cs:513.163,0)
--(axis cs:512.342,0)
--(axis cs:511.524,0)
--(axis cs:509.882,0)
--(axis cs:509.067,0)
--(axis cs:508.243,0)
--(axis cs:507.411,0)
--(axis cs:506.595,0)
--(axis cs:505.78,0)
--(axis cs:504.957,0)
--(axis cs:504.141,0)
--(axis cs:503.323,0)
--(axis cs:501.678,0)
--(axis cs:500.843,0)
--(axis cs:500.03,0)
--(axis cs:499.216,0)
--(axis cs:498.398,0)
--(axis cs:495.934,0)
--(axis cs:495.101,0)
--(axis cs:494.224,0)
--(axis cs:493.413,0)
--(axis cs:492.595,0)
--(axis cs:491.775,0)
--(axis cs:489.944,0)
--(axis cs:488.918,0)
--(axis cs:488.102,0)
--(axis cs:486.069,0)
--(axis cs:484.837,0)
--(axis cs:484.021,0)
--(axis cs:483.202,0)
--(axis cs:482.379,0)
--(axis cs:481.558,0)
--(axis cs:480.725,0)
--(axis cs:479.912,0)
--(axis cs:479.096,0)
--(axis cs:477.456,0)
--(axis cs:476.628,0)
--(axis cs:475.809,0)
--(axis cs:474.982,0)
--(axis cs:474.151,0)
--(axis cs:473.341,0)
--(axis cs:472.53,0)
--(axis cs:470.897,0)
--(axis cs:470.079,0)
--(axis cs:468.447,0)
--(axis cs:467.628,0)
--(axis cs:466.813,0)
--(axis cs:465.171,0)
--(axis cs:464.337,0)
--(axis cs:462.693,0)
--(axis cs:458.598,0)
--(axis cs:457.767,0)
--(axis cs:456.951,0)
--(axis cs:454.281,0)
--(axis cs:451.414,0)
--(axis cs:446.905,0)
--(axis cs:444.43,0)
--(axis cs:442.59,0)
--(axis cs:441.759,0)
--(axis cs:436.029,0)
--(axis cs:434.394,0)
--(axis cs:433.57,0)
--(axis cs:432.751,0)
--(axis cs:431.1,0)
--(axis cs:430.282,0)
--(axis cs:429.453,0)
--(axis cs:428.617,0)
--(axis cs:426.986,0)
--(axis cs:422.884,0)
--(axis cs:421.244,0)
--(axis cs:420.425,0)
--(axis cs:417.137,0)
--(axis cs:416.308,0)
--(axis cs:412.627,0)
--(axis cs:411.811,0)
--(axis cs:410.162,0)
--(axis cs:409.344,0)
--(axis cs:407.705,0)
--(axis cs:400.95,0)
--(axis cs:400.121,0)
--(axis cs:396.846,0)
--(axis cs:395.2,0)
--(axis cs:391.721,0)
--(axis cs:385.16,0)
--(axis cs:381.891,0)
--(axis cs:376.348,0)
--(axis cs:374.3,0)
--(axis cs:370.174,0)
--(axis cs:369.355,0)
--(axis cs:366.907,0)
--(axis cs:366.08,0)
--(axis cs:363.619,0)
--(axis cs:362.789,0)
--(axis cs:361.967,0)
--(axis cs:359.495,0)
--(axis cs:358.678,0)
--(axis cs:357.051,0)
--(axis cs:356.233,0)
--(axis cs:355.401,0)
--(axis cs:354.559,0)
--(axis cs:353.739,0)
--(axis cs:351.073,0)
--(axis cs:347.799,0)
--(axis cs:346.145,0)
--(axis cs:344.498,0)
--(axis cs:343.48,0)
--(axis cs:340.402,0)
--(axis cs:338.774,0)
--(axis cs:330.378,0)
--(axis cs:328.75,0)
--(axis cs:327.727,0)
--(axis cs:326.292,0)
--(axis cs:321.175,0)
--(axis cs:317.886,0)
--(axis cs:316.047,0)
--(axis cs:306.246,0)
--(axis cs:300.141,0)
--(axis cs:291.77,0)
--(axis cs:287.695,0)
--(axis cs:286.877,0)
--(axis cs:283.589,0)
--(axis cs:267.648,0)
--(axis cs:265.812,0)
--(axis cs:258.671,0)
--(axis cs:256.009,0)
--(axis cs:251.319,0)
--(axis cs:249.897,0)
--(axis cs:249.083,0)
--(axis cs:248.255,0)
--(axis cs:242.306,0)
--(axis cs:233.739,0)
--(axis cs:226.18,0)
--(axis cs:222.077,0)
--(axis cs:218.598,0)
--(axis cs:217.784,0)
--(axis cs:211.232,0)
--(axis cs:210.416,0)
--(axis cs:206.324,0)
--(axis cs:203.039,0)
--(axis cs:194.044,0)
--(axis cs:188.545,0)
--(axis cs:186.91,0)
--(axis cs:179.54,0)
--(axis cs:178.525,0)
--(axis cs:171.976,0)
--(axis cs:167.684,0)
--(axis cs:154.766,0)
--(axis cs:153.952,0)
--(axis cs:153.136,0)
--(axis cs:149.867,0)
--(axis cs:122.419,0)
--(axis cs:108.937,0)
--(axis cs:104.029,0)
--(axis cs:103.21,0)
--(axis cs:99.934,0)
--(axis cs:83.727,0)
--(axis cs:82.699,0)
--(axis cs:75.952,0)
--(axis cs:75.137,0)
--(axis cs:74.119,0)
--(axis cs:73.303,0)
--(axis cs:72.459,0)
--(axis cs:56.712,0)
--(axis cs:41.568,0)
--(axis cs:39.935,0)
--(axis cs:36.458,0)
--(axis cs:35.434,0)
--(axis cs:34.609,0)
--(axis cs:33.791,0)
--(axis cs:32.973,0)
--(axis cs:32.004,0)
--(axis cs:31.165,0)
--(axis cs:30.347,0)
--(axis cs:29.53,0)
--(axis cs:28.669,0)
--(axis cs:27.839,0)
--(axis cs:27.022,0)
--(axis cs:26.194,0)
--(axis cs:24.967,0)
--(axis cs:24.141,0)
--(axis cs:22.917,0)
--(axis cs:22.098,0)
--(axis cs:20.667,0)
--(axis cs:18.008,0)
--(axis cs:15.353,0)
--(axis cs:14.13,0)
--(axis cs:12.674,0)
--(axis cs:10.027,0)
--(axis cs:7.37,0)
--(axis cs:5.942,0)
--cycle;

\addplot [thick, green01270]
table {%
5.942 0.981
7.37 1.02028571428571
10.027 1.096375
12.674 1.20455555555556
14.13 1.2965
15.353 1.36690909090909
18.008 1.45672727272727
20.667 1.53263636363636
22.098 1.59918181818182
22.917 1.63609090909091
24.141 1.655
24.967 1.66009090909091
26.194 1.63090909090909
27.022 1.56945454545455
27.839 1.46872727272727
28.669 1.36481818181818
29.53 1.266
30.347 1.17754545454545
31.165 1.10681818181818
32.004 1.05472727272727
32.973 1.071
33.791 1.11345454545455
34.609 1.15872727272727
35.434 1.22454545454545
36.458 1.30336363636364
39.935 1.35936363636364
41.568 1.41272727272727
56.712 1.466
72.459 1.50881818181818
73.303 1.57636363636364
74.119 1.60954545454545
75.137 1.60763636363636
75.952 1.595
82.699 1.59463636363636
83.727 1.58118181818182
99.934 1.55327272727273
103.21 1.54345454545455
104.029 1.53545454545455
108.937 1.52690909090909
122.419 1.52927272727273
149.867 1.50627272727273
153.136 1.51072727272727
153.952 1.50763636363636
154.766 1.50827272727273
167.684 1.50463636363636
171.976 1.50509090909091
178.525 1.50027272727273
179.54 1.50709090909091
186.91 1.51545454545455
188.545 1.51845454545455
194.044 1.52363636363636
203.039 1.52309090909091
206.324 1.52236363636364
210.416 1.52081818181818
211.232 1.52163636363636
217.784 1.52145454545455
218.598 1.51872727272727
222.077 1.52554545454545
226.18 1.52036363636364
233.739 1.514
242.306 1.52127272727273
248.255 1.528
249.083 1.52563636363636
249.897 1.52763636363636
251.319 1.52527272727273
256.009 1.52454545454545
258.671 1.52909090909091
265.812 1.53209090909091
267.648 1.52863636363636
283.589 1.52963636363636
286.877 1.53109090909091
287.695 1.52572727272727
291.77 1.51518181818182
300.141 1.51854545454545
306.246 1.51936363636364
316.047 1.523
317.886 1.52790909090909
321.175 1.52409090909091
326.292 1.52545454545455
327.727 1.53063636363636
328.75 1.53054545454545
330.378 1.53490909090909
338.774 1.54236363636364
340.402 1.54463636363636
343.48 1.54736363636364
344.498 1.54845454545455
346.145 1.54981818181818
347.799 1.54754545454545
351.073 1.54736363636364
353.739 1.54309090909091
354.559 1.53881818181818
355.401 1.55218181818182
356.233 1.55627272727273
357.051 1.56090909090909
358.678 1.57090909090909
359.495 1.58690909090909
361.967 1.58254545454545
362.789 1.59872727272727
363.619 1.61472727272727
366.08 1.61336363636364
366.907 1.59763636363636
369.355 1.60818181818182
370.174 1.60718181818182
374.3 1.60136363636364
376.348 1.59418181818182
381.891 1.58409090909091
385.16 1.57672727272727
391.721 1.59063636363636
395.2 1.57436363636364
396.846 1.56072727272727
400.121 1.56927272727273
400.95 1.59563636363636
407.705 1.59536363636364
409.344 1.59354545454545
410.162 1.59345454545455
411.811 1.59027272727273
412.627 1.59354545454545
416.308 1.58236363636364
417.137 1.56945454545455
420.425 1.57036363636364
421.244 1.56954545454545
422.884 1.55327272727273
426.986 1.55936363636364
428.617 1.56754545454545
429.453 1.56918181818182
430.282 1.56690909090909
431.1 1.57654545454545
432.751 1.58409090909091
433.57 1.58336363636364
434.394 1.58954545454545
436.029 1.61663636363636
441.759 1.61145454545455
442.59 1.63290909090909
444.43 1.63618181818182
446.905 1.63045454545455
451.414 1.63336363636364
454.281 1.64618181818182
456.951 1.64818181818182
457.767 1.64327272727273
458.598 1.66927272727273
462.693 1.686
464.337 1.68109090909091
465.171 1.68781818181818
466.813 1.64881818181818
467.628 1.587
468.447 1.52463636363636
470.079 1.46718181818182
470.897 1.40690909090909
472.53 1.34790909090909
473.341 1.30954545454545
474.151 1.28554545454545
474.982 1.30272727272727
475.809 1.31945454545455
476.628 1.35227272727273
477.456 1.43536363636364
479.096 1.53254545454545
479.912 1.64990909090909
480.725 1.73436363636364
481.558 1.81763636363636
482.379 1.88409090909091
483.202 1.92563636363636
484.021 1.94009090909091
484.837 1.90736363636364
486.069 1.87845454545455
488.102 1.84345454545455
488.918 1.79945454545455
489.944 1.76254545454545
491.775 1.70063636363636
492.595 1.67745454545455
493.413 1.62481818181818
494.224 1.59136363636364
495.101 1.58436363636364
495.934 1.58527272727273
498.398 1.59627272727273
499.216 1.60681818181818
500.03 1.62254545454545
500.843 1.61536363636364
501.678 1.59672727272727
503.323 1.56245454545455
504.141 1.49709090909091
504.957 1.466
505.78 1.43409090909091
506.595 1.46490909090909
507.411 1.51618181818182
508.243 1.55609090909091
509.067 1.59154545454545
509.882 1.60590909090909
511.524 1.61836363636364
512.342 1.629
513.163 1.66554545454545
513.978 1.72463636363636
514.806 1.77818181818182
516.44 1.83154545454545
517.257 1.79754545454545
518.095 1.73781818181818
519.731 1.68981818181818
520.774 1.64309090909091
523.229 1.62118181818182
525.474 1.61318181818182
527.119 1.61309090909091
527.932 1.61745454545455
528.751 1.62518181818182
529.576 1.62754545454545
530.395 1.63372727272727
531.21 1.64363636363636
532.026 1.64427272727273
535.521 1.63527272727273
537.167 1.62645454545455
537.986 1.62336363636364
538.809 1.61981818181818
539.627 1.62418181818182
540.441 1.62045454545455
541.261 1.60990909090909
542.1 1.61163636363636
543.727 1.60718181818182
544.548 1.607
545.376 1.61863636363636
547.825 1.62327272727273
548.658 1.62918181818182
549.475 1.62745454545455
550.292 1.62927272727273
551.112 1.63118181818182
551.938 1.64027272727273
552.753 1.629
553.572 1.62872727272727
554.399 1.60281818181818
555.421 1.61254545454545
556.239 1.61081818181818
557.462 1.63109090909091
559.123 1.63972727272727
560.548 1.66090909090909
561.365 1.66590909090909
563.194 1.66345454545455
564.444 1.65
565.259 1.66218181818182
568.715 1.66109090909091
570.558 1.632
572.185 1.56263636363636
573.014 1.49845454545455
574.845 1.44627272727273
575.687 1.452
578.744 1.47227272727273
579.577 1.506
581.611 1.55081818181818
584.474 1.61263636363636
585.301 1.66727272727273
586.125 1.71872727272727
587.353 1.79772727272727
588.171 1.86081818181818
588.995 1.92127272727273
590.421 1.95336363636364
591.237 1.93690909090909
594.286 1.89154545454545
595.715 1.85
596.94 1.81363636363636
598.583 1.75054545454545
599.405 1.70536363636364
601.049 1.66663636363636
601.867 1.65
603.296 1.64818181818182
605.359 1.649
607.389 1.65972727272727
608.206 1.67118181818182
609.031 1.69209090909091
609.85 1.64618181818182
610.667 1.66772727272727
611.892 1.64227272727273
612.717 1.61872727272727
614.351 1.57627272727273
618.019 1.54427272727273
619.863 1.51263636363636
630.882 1.48109090909091
636.856 1.44418181818182
640.939 1.39609090909091
641.754 1.34527272727273
646.651 1.36590909090909
652.2 1.30954545454545
658.97 1.31327272727273
662.253 1.30318181818182
666.754 1.30872727272727
667.773 1.31781818181818
670.232 1.33036363636364
673.712 1.32236363636364
674.528 1.31436363636364
681.483 1.32663636363636
687.44 1.32718181818182
689.893 1.33063636363636
690.716 1.32645454545455
691.556 1.33018181818182
692.371 1.33463636363636
693.999 1.35327272727273
694.82 1.35236363636364
696.45 1.353
698.083 1.37345454545455
698.899 1.38181818181818
699.731 1.39427272727273
703.019 1.41818181818182
707.937 1.42954545454545
710.385 1.44045454545455
714.06 1.44990909090909
716.318 1.44381818181818
719.604 1.43209090909091
721.656 1.42781818181818
724.934 1.42527272727273
725.762 1.42281818181818
734.798 1.432
736.452 1.43054545454545
739.74 1.42718181818182
743.858 1.41545454545455
747.137 1.417
747.954 1.41445454545455
748.788 1.42209090909091
749.613 1.42618181818182
752.684 1.43127272727273
754.764 1.43
756.189 1.42281818181818
761.101 1.41845454545455
763.584 1.409
764.4 1.40836363636364
766.874 1.41718181818182
767.689 1.40772727272727
770.17 1.40354545454545
774.27 1.40145454545455
775.09 1.40036363636364
776.755 1.39672727272727
777.575 1.39436363636364
778.39 1.40218181818182
780.036 1.40436363636364
781.676 1.40918181818182
784.962 1.40681818181818
792.564 1.40654545454545
794.602 1.41309090909091
798.093 1.41190909090909
804.621 1.41509090909091
807.919 1.41381818181818
809.553 1.41390909090909
810.367 1.41590909090909
812.001 1.41081818181818
813.627 1.40918181818182
816.96 1.41436363636364
817.775 1.409
818.59 1.40590909090909
819.404 1.39481818181818
820.236 1.40327272727273
822.685 1.39418181818182
823.51 1.39481818181818
824.329 1.39281818181818
826.771 1.38909090909091
827.594 1.39072727272727
828.822 1.38790909090909
830.044 1.381
831.274 1.38190909090909
832.297 1.38018181818182
833.132 1.38927272727273
833.959 1.38272727272727
834.787 1.38518181818182
835.615 1.37790909090909
837.26 1.37454545454545
838.081 1.38036363636364
838.899 1.38181818181818
841.357 1.385
842.176 1.40190909090909
843.813 1.40854545454545
848.121 1.41381818181818
848.938 1.41672727272727
851.417 1.42409090909091
852.245 1.418
854.714 1.42036363636364
855.529 1.42290909090909
857.171 1.41263636363636
857.99 1.40936363636364
858.806 1.41245454545455
860.443 1.40218181818182
863.312 1.406
864.731 1.40518181818182
865.554 1.40545454545455
868.222 1.39818181818182
869.056 1.39490909090909
870.081 1.39272727272727
871.736 1.39309090909091
873.379 1.40554545454545
876.094 1.392
877.749 1.39481818181818
878.567 1.38945454545455
879.393 1.38436363636364
880.214 1.38381818181818
881.029 1.38654545454545
881.844 1.38445454545455
882.665 1.39581818181818
883.486 1.39890909090909
884.299 1.395
885.109 1.37181818181818
885.99 1.38727272727273
886.804 1.37263636363636
887.618 1.37218181818182
888.434 1.36827272727273
889.258 1.35381818181818
890.075 1.34581818181818
891.712 1.35036363636364
892.528 1.34909090909091
893.345 1.34681818181818
894.168 1.35
894.982 1.369
895.795 1.37181818181818
896.626 1.38318181818182
897.444 1.3828
898.26 1.38377777777778
899.08 1.398375
899.894 1.40285714285714
901.933 1.40183333333333
};
\addplot [purple, opacity=0.7, densely dashdotted, forget plot, very thick]
table {%
308.288 0.93195
308.288 2.05103181818182
};
\addplot [purple, opacity=0.7, densely dashdotted, forget plot, very thick]
table {%
605.356 0.93195
605.356 2.05103181818182
};

\node[fill=white, draw=black, anchor=south, font=\tiny, inner sep=1pt] at (axis cs:470,1.02) {Emergency};

\end{axis}

\end{tikzpicture}

%% file: figures/result2.tex
\begin{tikzpicture}[
    x=1.8mm, y=1.8mm,  
    >=latex,
    gnb_serving/.style={rectangle, fill=blue!60, draw=black, very thick, minimum size=10pt, inner sep=0pt, opacity=0.7},
    gnb_neighbor/.style={rectangle, fill=blue!60, draw=black, thick, minimum size=8pt, inner sep=0pt, opacity=0.7},
    ue_point/.style={circle, draw=black, thick, minimum size=12pt, inner sep=0pt, font=\bfseries\scriptsize},
    ue_retrain/.style={circle, draw=purple, very thick, line width=1.2pt, minimum size=12pt, inner sep=0pt, font=\bfseries\scriptsize},
]

\definecolor{detect0}{RGB}{165,0,38}       
\definecolor{detect17}{RGB}{230,80,50}     
\definecolor{detect25}{RGB}{244,109,67}    
\definecolor{detect50}{RGB}{255,255,191}   
\definecolor{detect75}{RGB}{166,217,106}   
\definecolor{detect83}{RGB}{102,189,99}    
\definecolor{detect85}{RGB}{90,180,95}     
\definecolor{detect100}{RGB}{0,104,55}     
\definecolor{alertzone}{RGB}{255,204,204}
\definecolor{alertedge}{RGB}{255,153,153}
\definecolor{neutralgray}{RGB}{176,176,176}
\colorlet{point2color}{detect83}
\colorlet{point4color}{detect17}
\colorlet{point5color}{detect50}
\colorlet{point6color}{detect85}
\colorlet{point8color}{detect100}
\colorlet{point9color}{detect100}

\begin{scope}
    \clip (-5, 0) rectangle (65, 39);
    \fill[alertzone, opacity=0.3] (10, 35) circle (14);
    \draw[alertedge, dashed, thick] (10, 35) circle (14);
\end{scope}
\fill[alertzone, opacity=0.3] (40, 15) circle (14);
\draw[alertedge, dashed, thick] (40, 15) circle (14);


\fill[red] (10, 37) -- (8, 33) -- (12, 33) -- cycle;
\draw[black, thick] (10, 37) -- (8, 33) -- (12, 33) -- cycle;
\node[font=\bfseries\scriptsize] at (10, 35) {D};
\fill[red] (40, 17) -- (38, 13) -- (42, 13) -- cycle;
\draw[black, thick] (40, 17) -- (38, 13) -- (42, 13) -- cycle;
\node[font=\bfseries\scriptsize] at (40, 15) {E};

\node[gnb_serving] at (25, 25) {};
\node[font=\bfseries\scriptsize, white] at (25, 25) {A};

\node[gnb_neighbor] at (40, 35) {};
\node[font=\bfseries\scriptsize, white] at (40, 35) {B};
\node[gnb_neighbor] at (10, 15) {};
\node[font=\bfseries\scriptsize, white] at (10, 15) {C};

\node[ue_point, fill=neutralgray] (p1) at (14.4, 14.6) {\textcolor{black}{1}};

\node[ue_point, fill=point2color] (p2) at (14, 32) {\textcolor{black}{2}};

\node[ue_point, fill=neutralgray] (p3) at (18, 19) {\textcolor{black}{3}};

\node[ue_point, fill=point4color] (p4) at (30, 10) {\textcolor{black}{4}};

\node[ue_retrain, fill=point5color] (p5) at (46, 5) {\textcolor{black}{5}};
\node[font=\bfseries\tiny, color=gray!60!black, align=center] at (46, 10) {Retraining\\triggered\\by LLM};

\node[ue_point, fill=point6color] (p6) at (38, 2.5) {\textcolor{black}{6}};
\node[font=\bfseries\tiny, color=gray!60!black] at (38, 0.5) {Retraining done};

\node[ue_point, fill=neutralgray] (p7) at (20, 7) {\textcolor{black}{7}};

\node[ue_point, fill=neutralgray] (p8) at (8, 18) {\textcolor{black}{8}};

\node[ue_point, fill=point8color] (p9) at (14, 38) {\textcolor{white}{9}};

\node[ue_point, fill=point9color] (p10) at (34, 6) {\textcolor{white}{10}};

\draw[->, gray!50!black, line width=1pt] (p1) -- (p2);  
\draw[->, gray!50!black, line width=1pt] (p2) to[out=0, in=90] (p3);  
\draw[->, gray!50!black, line width=1pt, dashed] (p3) -- (p4);  
\draw[->, gray!50!black, line width=1pt, dashed] (p4) -- (p5);  
\draw[->, gray!50!black, line width=1pt, dashed] (p5) to[out=-150, in=0] (p6);  
\draw[->, gray!50!black, line width=1pt, dashed] (p6) -- (p7);  
\draw[->, gray!50!black, line width=1pt] (p7) to[out=180, in=-45] (p8);  
\draw[->, gray!50!black, line width=1pt] (p8) -- (p9);  
\draw[gray!50!black, line width=1pt] (p9) to[out=-90, in=150] (17, 10);  
\draw[->, gray!50!black, line width=1pt, dashed] (17, 10) to[out=-30, in=150] (p10);  

\draw[->] (5, 0) -- (57, 0) node[right, font=\tiny, yshift=-2mm, xshift=-5mm] {X (m)};
\draw[->] (5, 0) -- (5, 39);
\node[rotate=90, font=\tiny] at (2.5, 17) {Y (m)};

\foreach \x/\l in {10/100, 20/200, 30/300, 40/400, 50/500} {
    \draw (\x, 0) -- (\x, -0.5) node[below, font=\tiny] {\l};
}
\foreach \y/\l in {0/0, 10/100, 20/200, 30/300} {
    \draw (5, \y) -- (4.5, \y) node[left, font=\tiny] {\l};
}

\fill[white, opacity=0.9] (17, 36.7) rectangle (56, 39);
\draw[black] (17, 36.7) rectangle (56, 39);

\fill[alertzone, opacity=0.5] (18, 37.3) rectangle (19, 38.3);
\draw[alertedge, dashed] (18, 37.3) rectangle (19, 38.3);
\node[right, font=\tiny] at (19.2, 37.8) {Interf.};

\fill[red] (25.4, 38.3) -- (24.4, 37.1) -- (26.4, 37.1) -- cycle;
\draw[black] (25.4, 38.3) -- (24.4, 37.1) -- (26.4, 37.1) -- cycle;
\node[right, font=\tiny] at (26.6, 37.8) {Other};

\node[gnb_serving, scale=0.4] at (32, 37.8) {};
\node[right, font=\tiny] at (32.8, 37.8) {Ours};

\draw[->, gray!50!black, line width=0.8pt] (37.5, 37.8) -- (39.5, 37.8);
\node[right, font=\tiny] at (39.7, 37.8) {Pre-tr.};
\draw[->, gray!50!black, line width=0.8pt, dashed] (44.5, 37.8) -- (46.5, 37.8);
\node[right, font=\tiny] at (46.7, 37.8) {Unseen};

\node[circle, fill=neutralgray, draw=black, minimum size=4pt, inner sep=0pt, font=\bfseries\tiny] at (52, 37.8) {t};
\node[right, font=\tiny] at (52.6, 37.8) {UE};

\fill[neutralgray] (58, 0) rectangle (59.2, 2.5);
\draw (58, 0) rectangle (59.2, 2.5);
\node[font=\scriptsize] at (61.5, 1.25) {N/A};
\shade[bottom color=detect0, top color=detect25] (58, 2.5) rectangle (59.2, 11.5);
\shade[bottom color=detect25, top color=detect50] (58, 11.5) rectangle (59.2, 20.5);
\shade[bottom color=detect50, top color=detect75] (58, 20.5) rectangle (59.2, 29.5);
\shade[bottom color=detect75, top color=detect100] (58, 29.5) rectangle (59.2, 38.5);
\draw (58, 2.5) rectangle (59.2, 38.5);
\foreach \p/\v in {2.5/0, 11.5/25, 20.5/50, 29.5/75, 38.5/100} {
    \draw (59.2, \p) -- (59.5, \p) node[right, font=\scriptsize] {\v};
}
\node[rotate=90, font=\scriptsize] at (63, 20.5) {Detection Rate (\%)};

\end{tikzpicture}